\documentclass{article}

\usepackage[preprint,nonatbib]{neurips_2024}

\usepackage[utf8]{inputenc} 
\usepackage[T1]{fontenc}    
\usepackage{hyperref}       
\usepackage{url}            
\usepackage{booktabs}       
\usepackage{amsfonts}       
\usepackage{nicefrac}       
\usepackage{microtype}      
\usepackage{graphicx}
\usepackage{cancel}
\usepackage{amsmath}
\usepackage{wrapfig}
\usepackage{threeparttable}
\usepackage{multirow}
\usepackage[table]{xcolor}
\usepackage{booktabs}
\usepackage{arydshln}
\usepackage{makecell}
\usepackage{algorithm}
\usepackage{algorithmic}
\usepackage{amssymb} 
\usepackage{float}
\usepackage{tabu}
\usepackage{subfig}

\newcommand{\tablestyle}[2]{\setlength{\tabcolsep}{#1}\renewcommand{\arraystretch}{#2}\centering\small}
\definecolor{lightgray}{gray}{0.8}
\definecolor{lightblue}{rgb}{0.21,0.49,0.74}
\hypersetup{
    colorlinks=true,
    breaklinks=true,
    urlcolor=lightblue,
    linkcolor=red,
    bookmarksopen=false,
    filecolor=black,
    citecolor=lightblue,
    linkbordercolor=lightblue
}

\def\ie{{\it{i.e.,~}}}

\vspace{-20mm}
\title{PLIP: Language-Image Pre-training for Person Representation Learning}

\author{%
	\textbf{Jialong Zuo $^{1}$} \quad ~\textbf{Jiahao Hong $^1$} \quad ~\textbf{Feng Zhang $^1$} \quad ~\textbf{Changqian Yu $^2$} \\ ~\textbf{Hanyu Zhou $^1$} \quad ~\textbf{Changxin Gao $^{1*}$} \quad ~\textbf{Nong Sang $^1$} \quad ~\textbf{Jingdong Wang $^3$}\\
    \\
	$^{1}$ National Key Laboratory of Multispectral Information Intelligent Processing Technology,\\ School of Artificial Intelligence and Automation, Huazhong University of Science and Technology, \\ $^{2}$ Skywork AI,  $^{3}$ Department of Computer Vision, Baidu Inc.   \\
    {\tt\small\{jlongzuo, cgao\}@hust.edu.cn, } {\tt\small\{wangjingdong\}@baidu.com}
}

\begin{document}

\maketitle

\vspace{-5mm}
\begin{abstract}
\vspace{-1mm}
Language-image pre-training is an effective technique for learning powerful representations in general domains. However, when directly turning to person representation learning, these general pre-training methods suffer from unsatisfactory performance. The reason is that they neglect critical person-related characteristics, i.e., fine-grained attributes and identities. To address this issue, we propose a novel language-image pre-training framework for person representation learning, termed PLIP. Specifically, we elaborately design three pretext tasks: 1) Text-guided Image Colorization, aims to establish the correspondence between the person-related image regions and the fine-grained color-part textual phrases. 2) Image-guided Attributes Prediction, aims to mine fine-grained attribute information of the person body in the image; and 3) Identity-based Vision-Language Contrast, aims to correlate the cross-modal representations at the identity level rather than the instance level. Moreover, to implement our pre-train framework, we construct a large-scale person dataset with image-text pairs named SYNTH-PEDES by automatically generating textual annotations. We pre-train PLIP on SYNTH-PEDES and evaluate our models by spanning downstream person-centric tasks. PLIP not only significantly improves existing methods on all these tasks, but also shows great ability in the zero-shot and domain generalization settings.
\end{abstract}

\vspace{-4mm}
\section{Introduction}
\vspace{-2mm}
Person-centric tasks, such as image/text-based person re-identification, person attribute recognition, person search and human parsing, are becoming increasingly influential in widespread applications, such as security monitor, smart city, virtual reality and scene understanding. Benefiting from the advances in designing task-specific methods~\cite{LGUR,ABDNet,rethinking,SeqNet,SCHP}, these tasks have achieved significant progress for the past few years. 

However, it has become evident through recent advancements in research that the mere development of sophisticated models based on specific tasks has already encountered a performance bottleneck. At the same time, some works~\cite{mae,mocov2,simclr,CAE} on general representation learning have shown great potential to further improve model performance. Due to the above reasons, some researchers~\cite{SOLIDER} have attempted to learn generic person representations by utilizing rich person images. However, their pre-training method based on sparse visual information falls short in terms of both training performance and efficiency.

Meanwhile, some works~\cite{CLIP,BLIP,lightningdot} have demonstrated that introducing the language modality helps to learn better representations in general domains, for the language naturally enjoys higher information density. However, when it comes to person representation learning, these general language-image pre-training methods like CLIP~\cite{CLIP} are often unsatisfactory in performance. We believe the reason is that they neglect some critical person-related characteristics, i.e., fine-grained attributes and identities. 

\begin{wrapfigure}[18]{r}{7cm} 
    \vspace{-0.5mm}
    \centering 
    \includegraphics[width=\linewidth]{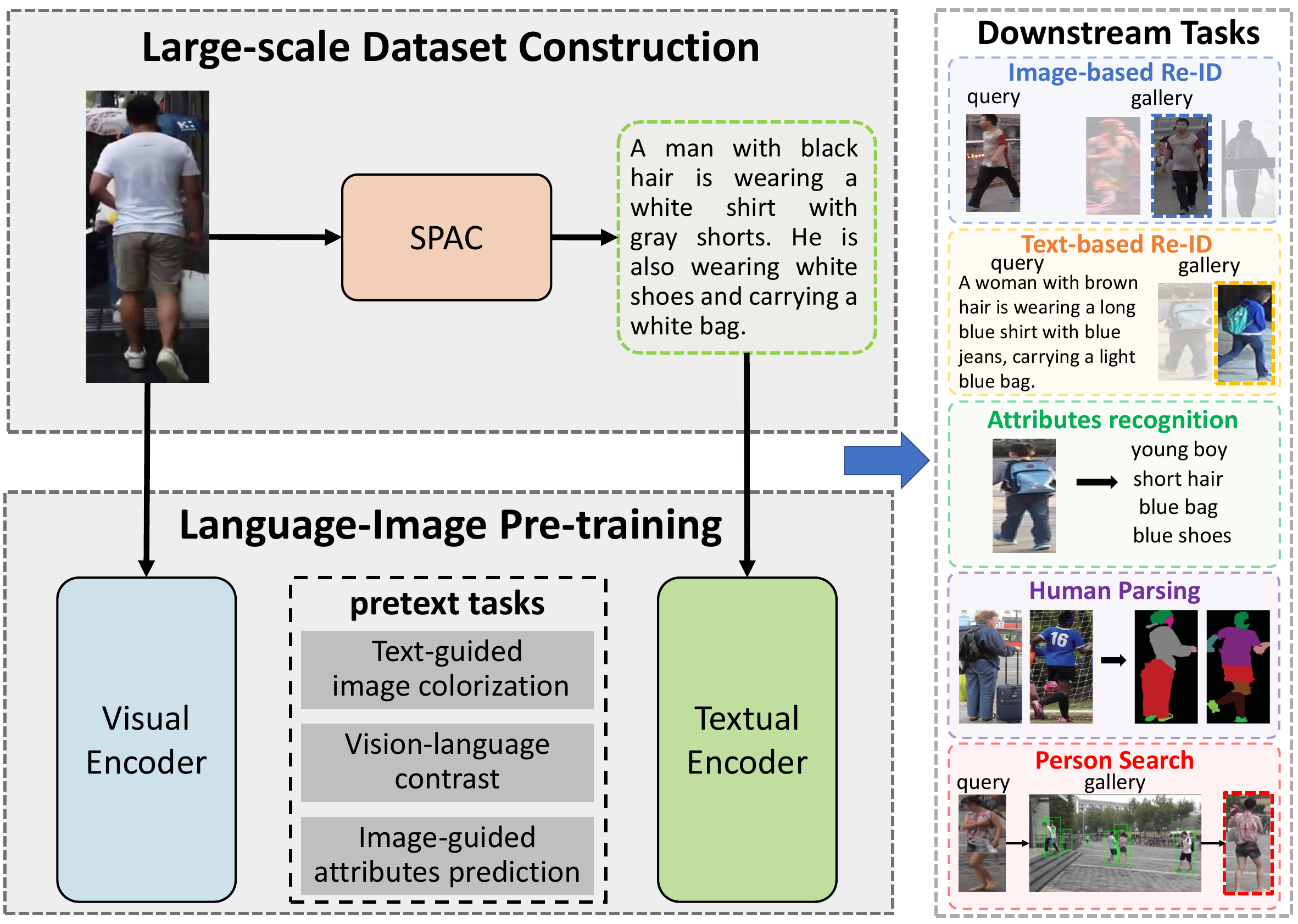}
    \caption{Illumination of our framework. Based on the constructed dataset, we pre-train a language-image model by three pretext tasks and transfer the model to some downstream person-centric tasks. 
    }
    \label{framework:fig1}
\end{wrapfigure}
On the one hand, these general language-image pre-training methods typically implement the global alignment between cross-modality representations and lack explicit consideration for fine-grained information. However, in person domain, the fine-grained information, such as the person attributes, plays a key role in distinguishing a specific person. Neglecting fine-grained attributes will easily lead to difficulty in learning discriminative person representations. On the other hand, they are based on instance level and only incorporate the concept of image-text pairs. They simply treat each image-text pair in the same person identity as different pairs, and assume that images and texts not in the same pair do not have a corresponding relationship. However, in person domain, there exists a notable concordance between the images and texts within the same person identity. If only conducting the optimization at the instance level, it will lead to instability to learn more meaningful representations.

To address these limitations mentioned above, we deeply consider the characteristics of persons and attempt to introduce the language modality into person representation learning. We propose a well-motivated language-image pre-training framework for learning generic and discriminative person representations, termed PLIP, to help the downstream person-centric tasks. Also, to implement the pre-training, we construct a large-scale person dataset with rich image-text pairs. The whole framework is illustrated in Fig.~\ref{framework:fig1}.

Specifically, to explicitly learn fine-grained and meaningful cross-modal associations, we design three pretext tasks in PLIP: 

(1)\textit{Text-guided Image Colorization}, given a complete textual description, is designed to restore the color information of a grayscale transformed person image. This task establishes the correspondence between the person-related image regions and the fine-grained color-part textual phrases, which robustly helps the model to learn the semantic concept of person body parts.

(2) \textit{Image-guided Attributes Prediction}, by exploiting the paired colorful images, is designed to predict the masked attribute phrases in textual descriptions. This task primarily focuses on predicting attributes, rather than predicting any random masked words as in the general domains. Through this multi-modal masked language modeling, it helps the model to understand the key areas and fine-grained attribute information of the person body in the image, which is crucial to identifying a person. 

(3) \textit{Identity-based Vision-language Contrast} is designed to associate representations between vision and language at the identity level rather than the instance level in general domains. This means it is optimized by narrowing the distance between any images and texts in the same person identity and widening the distance between those not in the same person identity. By taking identity into consideration, this task achieves more robust and meaningful association between different modalities.

As is well known, the quality and quantity of training data is essential for learning rich representations. However, there exists huge domain gap between the large-scale image-text pairs used in general domains and the specific person data. Also, the scale of the existing person datasets~\cite{textreid,SSAN} with manual textual descriptions is limited due to expensive hand-labeled annotations. Therefore, we construct a new large-scale person dataset with image-text pairs named SYNTH-PEDES based on the LUPerson-NL and LPW datasets~\cite{LUPnl,lpw}. The text annotations are automatically synthetized by our proposed person image captioner named Stylish Pedestrian Attributes-union Captioning (SPAC). The dataset contains 312,321 identities, 4,791,711 images and 12,138,157 textual descriptions. At the same time, extensive experiments have been conducted to verify the competitive quality of our synthetic dataset compared to manually annotated datasets~\cite{textreid,SSAN,RSTPReid}, which guarantees the superior performance of the learned person representations.

We utilize PLIP to pre-train our models on the SYNTH-PEDES dataset, and then evaluate the model performance on spanning downstream person-centric tasks. Extensive experiments show that our pre-trained models have learned generic person representations, pushing many state-of-the-art methods to a higher level on a wide range of person-centric tasks without bells and whistles. For example, for unsupervised person Re-ID, by applying our pre-trained ResNet50 model on PPLR~\cite{PPLR}, we improve the mAP metric on Market1501~\cite{market} and MSMT17~\cite{MSMT17} by 5.1\% and 14.7\%, respectively. The key contributions of this paper can be summarized as follows:

(1) We propose a novel language-image pre-training framework with three pretext tasks, termed PLIP, which deeply takes person-related characteristics into consideration. It can facilitate  fine-grained cross-modal association and learning generic person representations explicitly.

(2) To implement the pre-training, we construct a large-scale person dataset with generated text annotations, called SYNTH-PEDES. The dataset provides rich image-text pre-training data for this community.

(3) We pre-train PLIP on SYNTH-PEDES and the learned representations perform remarkable ability in various downstream person-centric tasks. It is demonstrated as generic to bring significant improvements to various baseline methods.

\vspace{-2mm}
\section{PLIP: Representation Learning Framework }
\vspace{-2mm}
This section presents our proposed language-image based person representation learning framework PLIP via three pretext tasks, \ie text-guided image colorization (TIC), image-guided attributes prediction (IAP), and identity-based vision-language contrast (IVLC). As illustrated in Fig.~\ref{fig:main}, the whole architecture is a dual branch encoder structure, and generic person representations can be learned through joint training of these three pretext tasks.

\vspace{-2mm}
\subsection{Text-guided Image Colorization}
\vspace{-1mm}
The TIC task is designed to restore the original color information of grayscale transformed images by exploiting the complete textual descriptions. Such cross-modal colorization promotes the construction of image-text association. The reason is that the attribute phrases in the descriptions generally contain fine-grained person-centric information especially color information and this colorization process naturally enables the model to understand the key components in textual descriptions and achieve a relationship construction between the textual phrases and visual regions. The overall task can be converted into a pixel-wise regression problem.

\begin{figure}[htb]
\centering
\includegraphics[width=\linewidth]{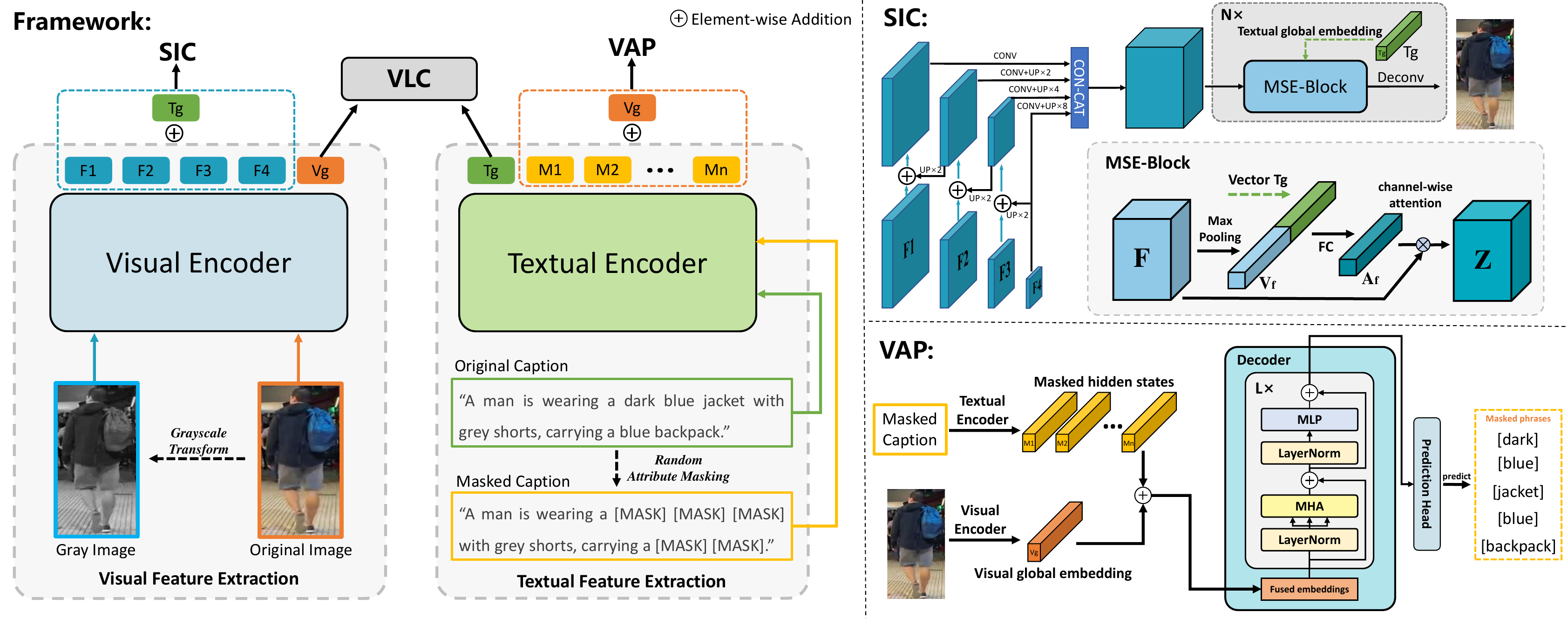}
\caption{Overview of our proposed framework incorporating a text-guided image colorization task, an image-guided attributes prediction task and an identity-based vision-language contrast task.}
\vspace{-5mm}
\label{fig:main}
\end{figure}

 As illustrated in Fig.~\ref{fig:main}, for a pair of a gray image and a complete textual description $\{\mathbf{i}_{gray},\mathbf{t}_{complete}\}$, in the encoding stage, the input gray image $\mathbf{i}_{gray}$ is firstly fed into a hierarchical network, which is as the visual encoder. Secondly, the hierarchical features are up-sampled to the same scale and concatenated to produce the feature $\mathbf{F}_{gray}$. Then, we feed the complete textual description $\mathbf{t}_{complete}$ to the textual encoder~\cite{bert} and adopts the average of the hidden-state outputs as the textual global embedding $\mathbf{T}_{global}$. 

In the decoding stage, the visual feature $\mathbf{F}_{gray}$ and textual global embedding $\mathbf{T}_{global}$ should be fused for colorization. Specifically, we adopt the multi-modal SE-blocks~\cite{Lapscore} as the cross-modal feature fusing module, so that the textual global embedding can play a role in the visual feature channels. In the block, the visual feature $\mathbf{F}_{visual}$ is compressed into a feature vector $\mathbf{V}_f$ through max pooling operation. Then we concatenate the visual feature vector and the textual global embedding $\mathbf{T}_{global}$, and feed the concatenated vector into several fc layers and a softmax layer to generate an attention vector $\mathbf{A}_f$. Finally, we utilize $\mathbf{A}_f$ on the visual feature $\mathbf{F}_{visual}$ to generate a multi-modal feature $\mathbf{Z}_{m}$. The decode is also made up of several de-convolution layers, which are employed to restore the feature dimensions. Finally, we generate a mutilmodal feature map with same dimensions as the input gray image $\mathbf{i}_{gray}$ and it can be utilized to predict the target color image $\mathbf{i}_{color}$.

 We denote $\theta_{tic}$ as the parameters of the trainable regression model mentioned above. It maps the textual global embedding $\mathbf{T}_{global}$ and the gray image extracted feature $\mathbf{F}_{gray}$ to the output recovered color image $\mathbf{i}_{color}$ as a target. TIC is supervised by:
\begin{equation}\label{eq7}
    \begin{aligned}
   \mathcal{L}_{tic} = \frac{1}{N}\sum^{N}\mathcal{D}(\mathbf{i}_{color},\theta_{tic}(\mathbf{T}_{global},\mathbf{F}_{gray})),
    \end{aligned}
\end{equation}
where $\mathcal{D}$ can be any differentiable distance function such as Euclidean distance we adopt and $N$ represents the total number of samples within a batch.

As displayed in Fig.~\ref{fig:sic_colorize} of the appendix, altering the color word in textual description significantly affects the colorization of image. 
However, our model may not fully understand the semantics of more detailed image regions. As shown in the last row, our model fails to distinguish between the blue clothing region and the red shoulder strap region (marked with yellow boxes), instead blending the two into a unified coloration.
This is due to the fact that the level of detail in manually annotated datasets is still not sufficient, resulting in the model being unable to theoretically learn representations with higher levels of detail and greater discrimination capabilities.
However, it is undeniable that our model has a preliminary understanding of the meaning of attributes and colors, and can associate them with related image regions, rather than simple memorization. This ability to distinguish between different parts of the person body guarantees the superior performance on the subsequent person-centric tasks. 

\vspace{-3mm}
\subsection{Image-guided Attributes Prediction}
The IAP task requires utilizing original color images to predict the masked attribute phrases in textual descriptions. Unlike previous methods~\cite{IRRA,Lapscore} that randomly mask any words or only color words in a description, our method focuses on masking attribute phrases. For each sentence, the attribute phrases are partially masked to create a masked textual description. In this multi-modal masked language modeling (MMLM) way, the correlation between images and texts can be bridged more in depth and more discriminative representations can be learned. The reason is that the prediction process enables the model to further understand the person-centric regions in images and extract the key semantic information. Meanwhile, by exploiting the visual representations, the MMLM enhances the perception of context and strengthens the interaction between vision and language modality.

 In the encoding stage, as illustrated in Fig.~\ref{fig:main}, for a pair of a color image $\mathbf{i}_{color}$ and a masked textual description $\mathbf{t}_{masked}$ with masked words $\mathbf{w}_m = \{\mathbf{w}_{m_{1}},\ldots,\mathbf{w}_{m_{M}}\}$ ($M$ is the number of masked words), we feed them to respective encoders to extract the visual global embedding $\mathbf{V}_{global}$ and textual hidden-state outputs $\mathbf{h}_t = \{\mathbf{h}_{t_{1}},\ldots,\mathbf{h}_{t_{L}}\}$ ($L$ is the length of the tokens in the masked textual description). The visual global embedding $\mathbf{V}_{global}$ is obtained from a pooling operation on the last stage feature output of the visual encoder. 
 
 In the decoding stage, to perform the IAP task, we specially design a simple but effective mutil-modal fusion module that mainly consists of self-attention blocks and a prediction head. Firstly, we adopt the element-wise summation of $\mathbf{h}_t$ and $\mathbf{V}_{global}$ as the preliminary fused embeddings. Then, the embeddings will be served as query($\mathcal{Q}$), key($\mathcal{K}$) and value($\mathcal{V}$) simultaneously. Finally, we obtain the multi-modal fused embeddings for each masked position by:

 \begin{equation}\label{eq8}
    \begin{aligned}
    \{\mathbf{h}_{m_{i}}\}_{i=1}^M = Blocks(\mathcal{Q},\mathcal{K},\mathcal{V}),
    \end{aligned}
\end{equation}
where $M$ is the total number of masked tokens in the masked textual description, and $Blocks$ are the self-attention blocks.

For each embedding representing the masked word, we use a prediction head to realize the corresponding probability prediction. The IAP can be optimized by minimizing the negative log-likelihood:

\begin{equation}\label{vap}
    \begin{aligned}
    \mathcal{L}_{iap} = - \frac{1}{MN}\sum^{N}\sum_{k=1}^{M}\log P(\mathbf{w}_{m_{k}}|\mathbf{h}_{m_{k}}),
    \end{aligned}
\end{equation}
where $M$ is the total number of masked words in a masked textual description, $N$ denotes the number of samples within a batch and $P$ denotes the probability distribution mapping.

\subsection{Identity-based Vision-language Contrast}
In order to further strengthen the correlation between vision and language modalities, we must optimize the model to learn a unified cross-modal feature space. A preliminary approach based on contrastive learning~\cite{CLIP} is to shorten the distance of the representations in image-text pairs and simultaneously amplify the distance of representations not in the same pair. This approach considers image-text pairs as instances and ignores the identity, where different image-text pairs of the same identity are treated as negative samples. However, in person-centric field, the identity plays a crucial role in distinguishing different people. Therefore, unlike the usual practices in general domain, we must take the identity into consideration.

For a group of a color image, a complete textual description and a corresponding identity $\{\mathbf{i}_{color},\mathbf{t}_{complete},Id\}$. The color image is firstly fed into the visual encoder. Then the last-stage feature is pooled to get the visual global embedding $\mathbf{V}_{global}$. The description is directly fed into the textual encoder and the average pooling of hidden-states will be served as the textual global embedding $\mathbf{T}_{global}$. 

Then, given a batch of $N$ image-text pairs, for each visual global embedding $\mathbf{V}^{i}_{global}$, we construct a set of visual-textual embedding pairs as $\{(\mathbf{V}^{i}_{global},\mathbf{T}^{j}_{global}),y_{i,j}\}^{N}_{j=1}$, where $y_{i,j}=1$ means that the pair is matched and from the same identity, and $y_{i,j}=0$ indicates the unmatched pair. Let $sim(\mathbf{x},\mathbf{z})=\mathbf{x}^{\top}\mathbf{z}/\|\mathbf{x}\|\|\mathbf{z}\|$ denotes the matching probability of $\mathbf{x}$ and $\mathbf{z}$. Then, the probability of matching pairs can be calculated with the following function:
\begin{equation}\label{eq555}
    \begin{aligned}
    p_{i, j}=\frac{\exp \left(sim\left(\mathbf{V}_{global}^{i}, \mathbf{T}_{global}^{j}\right) \right)}{\sum_{k=1}^{N} \exp \left(sim\left(\mathbf{V}_{global}^{i}, \mathbf{T}_{global}^{k}\right) \right)}.
    \end{aligned}
\end{equation}

Then, the IVLC loss from vision to language in a batch can be computed by:
\begin{equation}\label{eq9}
    \begin{aligned}
    \mathcal{L}_{v2l}=\frac{1}{N} \sum_{i=1}^{N} \sum_{j=1}^{N} p_{i, j} \log \left(\frac{p_{i, j}}{q_{i, j}+\epsilon}\right),
    \end{aligned}
\end{equation}
where $q_{i, j} =y_{i,j} / \sum_{k=1}^{N} y_{i,k}$ is the true matching probability and $\epsilon$ is a small number to avoid numerical problems.
    
Similarly, the IVLC loss from language to vision $\mathcal{L}_{l2v}$ can be
be computed by exchange $V_{global}$ and $T_{global}$ in above equations. Finally, the IVLC task can be optimized by:

\begin{equation}\label{eq10}
    \begin{aligned}
    \mathcal{L}_{IVLC}=\mathcal{L}_{v2l}+\mathcal{L}_{l2v},
    \end{aligned}
\end{equation}

Indeed, the IVLC loss can be any cross-modal contrastive loss that takes identity into consideration such as the Cross-Modal Projection Matching (CMPM) loss~\cite{CMPM} we adopt, which promotes the representation association between multiple modalities by incorporating the cross-projection into KL divergence.

Then, to supervise the model to learn discriminative and highly generic person representations, the overall multi-task loss $\mathcal{L}$ is computed as:
\begin{equation}\label{eq11}
    \begin{aligned}
    \mathcal{L} = \mathcal{L}_{ivlc}+\lambda_{1}\mathcal{L}_{tic}+\lambda_{2}\mathcal{L}_{iap},
    \end{aligned}
\end{equation}
where $\lambda_1,\lambda_2 \in \mathbb{R}^{+}$ are hyper-parameters to control the importance of each pretext task.

\section{SYNTH-PEDES: A Large-scale Image-text Person Dataset}
\begin{figure}[htb]
\centering
\includegraphics[width=0.95\linewidth]{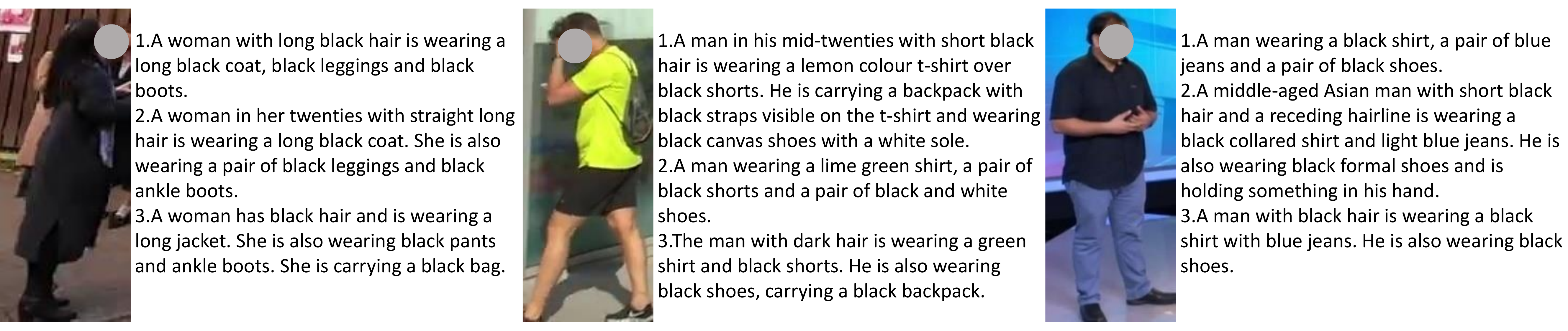}
\caption{Visualization of some examples in our SYNTH-PEDES dataset.}
\label{fig:captionsamples}
\vspace{-5mm}
\end{figure}

\vspace{-2mm}
We build the SYNTH-PEDES dataset to pre-train our PLIP models at a large-scale. In this section, we show the general process of constructing our SYNTH-PEDES dataset, which can be described as three steps. The complete construction details can be found in Sec.~\ref{howtoconstructsynthpedes} of the appendix.

Firstly, we collect and process two large-scale person datasets to form the image dataset. The first is LUPerson-NL~\cite{LUPnl}. It is a new variant of LUPerson~\cite{LUP} on top of raw videos from LUPerson and assign the noisy labels to each person image with automatically generated tracklet. It consists of 10\textit{M} images with about 430\textit{K} identities collected from 21\textit{K} scenes. The second is LPW~\cite{lpw}. It consists of 2,731 different persons and 592,438 images collected from three different crowded scenes. 

Secondly, we propose an image captioner named SPAC to generate satisfactory textual descriptions for the images. Given an input person image, there is no specific work targeting at generating captions that detailedly describe the person's appearance. To this end, we propose a simple but effective method for person image captioning. It can generate attribute-annotations and stylish textual-descriptions, which simulate the diverse perspectives that different annotators may have on the same person image. The specific technics for SPAC can be found in Sec.~\ref{datasetconstructdetail} of the appendix.

Thirdly, we adopt some post-processing approaches to eliminate the noises and improve the dataset quality. We propose Seed Filter Strategy to filter the noises in LUPerson-NL, which includes three processes of Filter-out, Reassignment and Merger. Meanwhile, we propose Data Distribution Strategy to ensure the quality of generated attributes, the consistency of gender annotation, and identity distribution balance. The specific details for the strategies can be found in Sec.~\ref{Strategydetail} of the appendix.

Thanks to the outstanding generating ability of our proposed SPAC, the SYNTH-PEDES dataset is full of high-quality textual descriptions in a variety of styles, which can be utilized to train the representation learning model. Compared with existing person datasets, SYNTH-PEDES has the following advantages: 

\noindent
\textbf{Diversified.} Our dataset contains a wide range of variations in the textual descriptions. Unlike the previous person datasets with only one or two image-text pairs, most images of our dataset are annotated with three textual descriptions. 

\noindent
\textbf{High-quality.} As some typical qualitative examples can be seen in Fig.~\ref{fig:captionsamples}, the generated annotations achieve an accurate and detailed description of the person appearance. The further experiments conducted on the dataset quality evaluation can be found in Sec.~\ref{datasetqualityeval} of the appendix. Researchers can use this dataset with confidence to conduct relevant studies. 

\noindent
\textbf{Large-scale.} In Tab.~\ref{table1} of the appendix, we have compared the properties of SYNTH-PEDES with other popular person datasets. As we can see, SYNTH-PEDES is the largest real person dataset with high-quality image-text pairs by far, which contains 312,321 identities, 4,791,711 images, and 12,138,157 textual descriptions. 

\vspace{-3mm}
\section{Experiments}
\vspace{-2mm}
\noindent
\textbf{Implementation}. During the training of PLIP, we adopt four types of backbone as the visual encoder, \ie ResNet50, ResNet101, ResNet152 and Swin Transformer Base. The pre-trained BERT~\cite{bert} is utilized as the textual encoder with the last 5 layers unfrozen. We train our model on 4 × Geforce 3090 GPUs for 70 epochs. For each person-centric downstream task, we reproduce a range of state-of-the art methods as the baselines. If not specially stated, we perform the experiments by just replacing the backbone in each baseline to the our pre-trained models. We perform in-depth experiments on eleven datasets for five downstream person-centric tasks. Meanwhile, we perform thorough ablation studies and analyses in Sec.~\ref{throughablation} of the appendix. More details and experiments can be found in the appendix. 

\begin{table}[t]
\tiny
\centering
\caption{Performance comparisons of our PLIP with some other SoTA pre-trained models on downstream tasks. The table is divided into two parts: the upper part covers some general-domain pre-trained models and the lower part focuses on some person-domain pre-trained models. The baseline downstream methods for the five tasks are CMPM/C~\cite{CMPM}, ABDNet~\cite{ABDNet}, Rethinking~\cite{rethinking}, SeqNet~\cite{SeqNet} and SCHP~\cite{SCHP}, respectively. Due to the inability of certain non-hierarchical models to be applied to some downstream methods, their performance results are indicated as ``-".}
\label{comparisonwithothers}
\resizebox{\linewidth}{!}{
\begin{tabular}{cc|cc|cc|cc|cc|cc}
\hline
\multicolumn{1}{c}{\multirow{2}{*}{Method}} &\multicolumn{1}{c|}{\multirow{2}{*}{Backbone}} & \multicolumn{2}{c|}{Text Re-ID}&\multicolumn{2}{c|}{Image Re-ID} &\multicolumn{2}{c|}{Attribute Recog.}&\multicolumn{2}{c|}{Person Search}&\multicolumn{2}{c}{Human Parsing}\\

 & & CUHK & ICFG & Market & Duke & PETA & PA100K & SYSU & PRW & LIP &PASCAL  \\

\hline
Baseline&RN50 &54.81/83.22 &47.61/75.48 &88.3/95.6 &78.6/89.0 &83.96/86.35 &80.21/87.40 &94.8/95.7 &47.6/87.6 &59.36 &71.46\\
Baseline&RN101 &56.77/84.86 &51.78/77.92 &89.3/95.7 &79.8/88.9 &85.17/86.44 &81.61/87.62&95.1/96.1 &48.4/87.7  &60.57 &72.12\\
Baseline&RN152 &58.64/85.84 &53.13/79.18 &89.6/95.8 &80.1/89.0 &85.81/86.64 &82.17/87.96 &95.3/96.3 &48.8/87.9 &61.71 &72.87\\
\hdashline[2.5pt/5pt]

MoCov3~\cite{mocov3}&RN50 &52.17/81.94 &46.51/74.86 &87.7/95.0 &77.3/87.7 &82.87/85.19 &79.12/86.56 &94.1/94.5 &46.2/86.1 &57.63 &69.16\\
MoCov3~\cite{mocov3}&ViT-S &53.21/82.63 &46.77/74.62 &-/- &-/- &83.51/85.98 &79.64/87.12 &-/- &-/- &- &-\\
MoCov3~\cite{mocov3}&ViT-B &55.46/83.82 &50.28/76.22 &-/- &-/- &84.82/86.83 &80.96/87.28 &-/- &-/- &- &-\\

\hdashline[2.5pt/5pt]
CLIP~\cite{CLIP}&RN50 &57.92/85.12 &51.30/77.64 &89.5/95.7 &79.7/88.9 &84.49/86.65 &81.07/87.82 &92.8/93.4 &48.0/86.7 &59.81 &70.93\\
CLIP~\cite{CLIP}&RN101 &58.04/85.88 &52.80/79.31 &90.6/96.0&81.4/89.9 &85.67/87.04 &81.82/87.95 &93.3/93.8 &48.5/87.4 &60.74 &72.39\\
CLIP~\cite{CLIP}&ViT-B &58.91/86.21 &53.73/79.98 &-/- &-/- &86.37/87.95 &82.41/88.41 &-/- &-/-&- &-\\
\hdashline[2.5pt/5pt]

BLIP~\cite{BLIP}&ViT-B&60.37/86.63 &54.64/80.23 &-/- &-/- &86.97/88.42 &82.81/88.46 &-/- &-/- &- &-\\
BLIP~\cite{BLIP}&ViT-L &63.23/88.74 &56.96/82.79 &-/- &-/- &87.08/88.55 &83.23/88.62 &-/- &-/- &- &-\\

\hline
LUP~\cite{LUP}&RN50 &57.51/84.86 &50.52/76.63 &90.2/95.9 &80.7/89.3 &84.17/86.10&81.94/88.25 &94.3/94.9 &47.7/87.3 &59.43 &71.89\\
LUP~\cite{LUP}&RN101 &57.84/85.36 &53.08/78.93 &91.4/96.5 &83.1/90.9 &85.66/86.57 &82.31/88.21 &94.8/95.5 &47.9/87.4 &60.41 &72.35\\
LUP~\cite{LUP}&RN152 &59.41/86.22 &53.79/79.51 &91.7/96.7 &83.7/91.9 &85.92/87.44 &82.63/88.47 &95.3/95.9 &48.6/88.1 &62.05 &73.16\\
\hdashline[2.5pt/5pt]

LUP-NL~\cite{LUPnl}&RN50 &57.85/84.97 &50.64/76.55 &90.8/96.1 &81.7/89.7 &84.09/86.32 &81.71/88.08 &95.4/95.9 &49.3/87.5 &59.88 &71.67\\
LUP-NL~\cite{LUPnl}&RN101 &58.12/85.74 &53.38/78.85 &91.4/96.5 &82.1/90.9 &85.77/86.93 &82.39/88.28 &95.5/95.7 &49.6/87.6 &60.92 &72.51\\
LUP-NL~\cite{LUPnl}&RN152 &59.27/86.06 &54.21/80.12 &91.5/96.5 &81.8/90.6 &86.18/88.09 &82.54/88.65 &95.8/96.1 &49.5/88.2 &61.98 &73.12\\
\hdashline[2.5pt/5pt]

SOLIDER~\cite{SOLIDER}&Swin-T &51.22/80.27 &46.51/74.16 &87.2/94.6 &77.5/87.8 &84.12/86.27 &83.81/87.91 &94.9/95.7 &56.8/86.8 &57.52 &68.42\\
SOLIDER~\cite{SOLIDER}&Swin-S &56.43/84.78 &51.19/77.82 &89.6/95.5 &80.2/89.3 &85.47/87.14 &85.32/89.08 &95.5/95.8 &\textbf{59.8}/86.7 &60.21 &69.81\\
SOLIDER~\cite{SOLIDER}&Swin-B &58.06/85.64 &52.37/79.62 &90.1/95.6 &81.2/89.8 &85.78/87.46 &\textbf{85.91}/\textbf{89.66} &94.9/95.5 &59.7/86.8 &60.50 &70.02\\

\hdashline[2.5pt/5pt]

PLIP(ours)&RN50 &71.13/92.81 &64.88/87.32 &91.4/96.8 &81.7/91.1 &85.12/86.94 &82.21/88.62 &96.0/96.7 &53.5/88.9 &60.41 &72.14\\
PLIP(ours)&RN101 &72.34/93.46 &64.92/87.66 &92.0/96.9 &82.3/91.8 &85.84/87.42&82.52/88.16 &96.2/96.8 &54.6/89.4 &61.32 &72.63\\
PLIP(ours)&RN152 &73.68/94.22 &65.52/88.40 &92.6/97.1 &83.1/92.1 &86.28/88.17 &82.91/88.83 &\textbf{96.5}/\textbf{97.0} &55.7/89.8 &62.44 &73.51\\
 \rowcolor{gray!20}
PLIP(ours)&Swin-B &\textbf{75.36}/\textbf{94.87} &\textbf{66.17}/\textbf{88.94} &\textbf{93.2}/\textbf{97.3} &\textbf{84.4}/\textbf{92.2} &\textbf{87.12}/\textbf{88.84} &83.65/89.17 &96.4/96.7 &56.6/\textbf{89.9} &\textbf{63.52} &\textbf{73.93}\\
\hline
\end{tabular}}
\vspace{-4mm}
\end{table}

\subsection{Comparison With Other Pre-trained Models}
\vspace{-2mm}
We have compared the performance of our PLIP pre-trained models with other SoTA pre-trained models~\cite{mocov3,LUP,LUPnl,SOLIDER,CLIP,BLIP,APTM} on five downstream tasks. We reproduce a range of popular and open-sourced methods~\cite{CMPM,ABDNet,rethinking,SeqNet,SCHP} as the baselines, which are initialized by ImageNet supervised pre-trained ResNet backbones. Then, we compare the performance of different pre-trained models by simply replacing the backbone in each method. The compared models can be divided into general-domain pre-trained models and person-domain pre-trained models, with the latter typically exhibiting better performance in person-centric tasks. The performance metrics for these tasks are R@1/R@10, mAP/R@1, mAP/F1, mAP/R@1 and mIoU, respectively. As shown in Tab.~\ref{comparisonwithothers}, our PLIP models consistently demonstrate a significant performance advantage over other pre-trained models in a fair comparison with roughly equal parameters. 

\vspace{-4mm}
\subsection{Evaluation on Text-based Person Re-ID}
\vspace{-2mm}
\begin{wraptable}[17]{r}{7cm}
\vspace{-5mm}
\tiny
\tabcolsep=3.2pt
\caption{The results of our transfer experiments. We show the best score in bold. \textit{z-s}: zero-shot setting; \textit{l-p}: linear-probing setting; \textit{f-t}: fine-tune setting. $\ddagger$ stands for the results reproduced with public checkpoints released by the authors.}
\label{tab1}
\resizebox{\linewidth}{!}{
\begin{tabular}{c|ccc|ccc}
\hline
\multicolumn{1}{c|}{\multirow{2}{*}{Method}} & \multicolumn{3}{c|}{CUHK-PEDES}&\multicolumn{3}{c}{ICFG-PEDES}\\
\cline{2-7}
 & R@1 & R@5 & R@10 & R@1 & R@5 & R@10  \\
\hline
\multicolumn{1}{c|}{ViTAA~\cite{vitaa}} &55.97 &75.84 &83.52&50.98 &68.79 &75.78\\

\multicolumn{1}{c|}{SSAN~\cite{SSAN}} &61.37& 80.15 &86.73	&54.23 &72.63 &79.53\\

\multicolumn{1}{c|}{LapsCore~\cite{Lapscore}} &63.40& -& 87.80	&- &- &-\\

\multicolumn{1}{c|}{TIPCB$\ddagger$~\cite{tipcb}} &63.63& 82.82& 89.01&54.96 &74.72 &81.89\\ 

\multicolumn{1}{c|}{LGUR~\cite{LGUR}} &64.21& 81.94& 87.93&57.42 &74.97 &81.45\\

\hline
\multicolumn{1}{c|}{PLIP+\textit{z-s}} &52.86& 74.69& 82.46	&49.86 &68.78 &76.27\\
\multicolumn{1}{c|}{PLIP+\textit{l-p}} &63.63&82.85& 89.36	&58.51 &77.83 &84.24\\\rowcolor{gray!20}
\multicolumn{1}{c|}{PLIP+\textit{f-t}} &\textbf{70.11}& \textbf{86.60}& \textbf{91.89}	&\textbf{64.58} &\textbf{81.30} &\textbf{86.82}\\
\hline
\end{tabular}}
\end{wraptable}
\textbf{Transfer capability.} To evaluate the transfer capability of our pre-trained models, we conduct three different experiments with ResNet50 as the visual encoder. Firstly, we directly evaluate the model's zero-shot performance without any extra fine-tuning. Secondly, we perform linear probing by adding a trainable linear embedding layer to each modal frozen encoder. Finally, we unfreeze all encoders and perform fine-tuning. We use the simple CMPM loss~\cite{CMPM} as the training target. From Tab.~\ref{tab1} we can see, our pre-trained model is not only competitive with some fully supervised methods~\cite{vitaa,SSAN} even without fine-tune training, but also greatly exceeds them by a large margin with a simple fine-tune. These results demonstrate that our pre-trained models have excellent transfer capability for this task.

\begin{wraptable}[6]{r}{7cm}
\vspace{-19mm}
\tiny
\tabcolsep=3.2pt
\caption{Comparison on domain generalization. ``C'' and ``I'' denote CUHK-PEDES and ICFG-PEDES, respectively.}
\vspace{-2mm}
\label{tab2}
\resizebox{\linewidth}{!}{
\begin{tabular}{c|ccc|ccc}
\hline
\multicolumn{1}{c|}{\multirow{2}{*}{Method}} & \multicolumn{3}{c|}{$C {\rightarrow}I$}&\multicolumn{3}{c}{$I{\rightarrow}C$}\\
\cline{2-7}
 & R@1 & R@5 & R@10 & R@1 & R@5 & R@10  \\
\hline
\multicolumn{1}{c|}{Dual Path~\cite{dualpath}} &15.41 &29.80 &38.19&7.63 &17.14 &23.52\\

\multicolumn{1}{c|}{SCAN~\cite{scan}} &21.27& 39.26&48.83&13.63 &28.61 &37.05\\

\multicolumn{1}{c|}{SSAN~\cite{SSAN}} &29.24& 49.00& 58.53	&21.07 &38.94&48.54\\

\multicolumn{1}{c|}{LGUR~\cite{LGUR}} &34.25& 52.58& 60.85	&25.44 &44.48&54.39\\

RaSa~\cite{RASA} &50.59&67.46&74.09&50.70&72.40&79.58\\
\rowcolor{gray!20}
\hline
\multicolumn{1}{c|}{PLIP} &\textbf{56.64}& \textbf{75.65}&\textbf{82.38}	&\textbf{57.34} &\textbf{77.60} &\textbf{84.49}\\
\hline
\end{tabular}}
\end{wraptable}

\textbf{Domain generalization.} To verify our models' domain generalization capability, we carry out experiments with cross-domain settings. We use the CMPM loss as the training target. As illustrated in Tab.~\ref{tab2}, our model achieves improvements by large margins when compared with all other methods. Specifically, our model outperforms LGUR by 22.4\% and 31.9\% in terms of Rank-1 metric on the $C{\rightarrow}I$ and $I{\rightarrow}C$ settings, respectively. These results demonstrate that our pre-trained models have great capability in domain generalization for this task.

\begin{wraptable}[15]{r}{7cm}
\vspace{-5mm}
\tiny
\tabcolsep=3.2pt
\centering
\caption{Comparison on three methods by using different pre-trained models. All results are shown in Rank-1/Rank-10. }
\vspace{-1mm}
\label{tab3}
\resizebox{\linewidth}{!}{
\begin{tabular}{c|c|ccc}
\hline
\multicolumn{1}{c|}{}&\multicolumn{1}{c|}{Pre-train} & CMPM/C~\cite{CMPM} & SSAN~\cite{SSAN} & LGUR~\cite{LGUR}\\
\hline
\multicolumn{1}{c|}{\multirow{6}{*}{\rotatebox{90}{CUHK-PEDES}}}
&{Baseline.} &54.81/83.22 &61.37/86.73& 64.21/87.93\\
&{MoCov3} &52.17/81.94& 61.97/86.63& 65.33/88.47\\
&{CLIP} &57.92/85.12& 62.09/86.89& 64.70/88.76\\
&{LUP} &57.51/84.86& 63.91/88.36& 65.42/89.36\\
&{LUP-NL} &57.85/84.97& 63.71/87.46 &64.68/88.69 \\

&\cellcolor{gray!20}{PLIP} &\cellcolor{gray!20}\textbf{71.13/92.81}& \cellcolor{gray!20}\textbf{65.90/89.44}& \cellcolor{gray!20}\textbf{67.62/89.90}\\

\hline
\multicolumn{1}{c|}{\multirow{6}{*}{\rotatebox{90}{ICFG-PEDES}}}
&{Baseline.} &47.61/75.48 &54.23/79.53&57.42/81.45 \\
&{MoCov3} &46.51/74.86& 55.27/79.64& 59.90/82.94 \\
&{CLIP} &51.30/77.64& 53.58/78.96& 58.35/82.02\\
&{LUP} &50.52/76.63 &56.51/80.41& 60.33/83.06\\
&{LUP-NL} &50.64/76.55& 55.59/79.78 &60.25/82.84\\

&\cellcolor{gray!20}{PLIP} &\cellcolor{gray!20}\textbf{64.88/87.32}& \cellcolor{gray!20}\textbf{60.70/83.17}& \cellcolor{gray!20}\textbf{62.38/84.28}\\
\hline
\end{tabular}}
\vspace{-5mm}
\end{wraptable}
\textbf{Improvement over existing methods.} We reproduce three representative baseline methods~\cite{CMPM,SSAN,LGUR} and explore the performance difference by changing the encoders with different pre-trained models. From Tab.~\ref{tab3}, we can see that equipped with our pre-trained model, all the baseline methods achieve higher and best accuracy on each dataset. It is worth noting that, due to the fact that our PLIP models have already learned an excellent joint visual-textual feature space through large-scale pre-training, achieving outstanding performance is easily attained by fine-tuning with the simple CMPM/C~\cite{CMPM} loss rather than complicated designing such as SSAN~\cite{SSAN} and LGUR~\cite{LGUR}.

\begin{wraptable}[19]{r}{7cm}
\vspace{-7mm}
\tabcolsep=3.2pt
\centering
\caption{Comparison with the state-of-the-art methods on text-based person Re-ID. We show the best score in bold.}
\vspace{-2mm}
\label{tab4}
\resizebox{\linewidth}{!}{
\begin{tabular}{c|cc|ccc|ccc}
\hline
\multicolumn{1}{c|}{\multirow{2}{*}{Method}} & \multicolumn{1}{c}{\multirow{2}{*}{Image }}&\multicolumn{1}{c|}{\multirow{2}{*}{Text }}&\multicolumn{3}{c|}{CUHK-PEDES}&\multicolumn{3}{c}{ICFG-PEDES}\\
\cline{4-9}
& & & R@1 & R@5 & R@10 & R@1 & R@5 & R@10  \\
\hline
\multicolumn{1}{c|}{GNA-RNN~\cite{textreid}}&VGG & LSTM&19.05 &- &53.64&- &- &-\\

\multicolumn{1}{c|}{CMPM/C~\cite{CMPM}}&RN50 &LSTM&49.37& -& 79.27	&43.51 &65.44 &74.26\\

\multicolumn{1}{c|}{ViTAA~\cite{vitaa}}&RN50 &LSTM &55.97&75.84 &83.52&50.98&68.79&75.78\\

\multicolumn{1}{c|}{NAFS~\cite{NAFS}}&RN50 &BERT &59.94& 79.86& 86.70&-&-&-\\

\multicolumn{1}{c|}{SSAN~\cite{SSAN}}&RN50 &LSTM &61.37 &80.15 &86.73&54.23&72.63&79.53\\

\multicolumn{1}{c|}{LapsCore~\cite{Lapscore}}&RN50 &BERT &63.40 &- &87.80&-&-&-\\

\multicolumn{1}{c|}{TIPCB$\ddagger$~\cite{tipcb}} &RN50 &BERT&63.63& 82.82& \underline{89.01}&54.96 &74.72 &81.89\\ 

\multicolumn{1}{c|}{LGUR~\cite{LGUR}}&RN50 &BERT &64.21 &81.94 &87.93&57.42 &74.97 &81.45\\ 

\hdashline[2.5pt/5pt]
\multicolumn{1}{c|}{IVT~\cite{IVT}}&ViT-B &BERT &65.59 &83.11 &89.21&56.04 &73.60 &80.22\\

\multicolumn{1}{c|}{CFine~\cite{CFine}}&ViT-B &BERT &69.57 &85.93 &91.15&60.83 &76.55 &82.42\\

\multicolumn{1}{c|}{IRRA~\cite{IRRA}}&ViT-B & Xformer&73.38 &89.93 &93.71&63.46 &80.25 &85.82\\

\multicolumn{1}{c|}{APTM~\cite{APTM}}&Swin-B & BERT&76.17&89.47&93.57&\textbf{68.22} &82.87 &87.50\\

\multicolumn{1}{c|}{RaSa~\cite{RASA}}&ViT-B & BERT&\textbf{76.51}&90.29&94.25&65.28 &80.40 &85.12\\

\hline

\multicolumn{1}{c|}{PLIP}&RN50 &BERT &71.13  &87.57  &92.81	&64.88 &81.70&  87.32\\
\multicolumn{1}{c|}{PLIP}&RN101 &BERT &72.34  &89.15  &93.46	&64.92 &81.63&  87.66\\
\multicolumn{1}{c|}{PLIP}&RN152 &BERT &73.68  &90.35  &94.22	&65.52 &82.64&  88.40\\

 \rowcolor{gray!20}
\multicolumn{1}{c|}{PLIP}&Swin-B &BERT &75.36  &\textbf{90.86}  &\textbf{94.87}	&66.17 &\textbf{83.37}&  \textbf{88.94}\\

\hline
\end{tabular}}
\end{wraptable}
\textbf{Comparison with state-of-the-art methods.} In Tab.~\ref{tab4}, we compare our results with some SoTA methods on each dataset. 
The compared methods can be classified based on whether they rely on the multi-modal pre-trained models. Generally, multi-modal pre-trained models can bring about noticeable performance improvements for this task. 
It is worth noting that RaSa utilizes a larger image resolution of 384×384, while APTM adopts a two-stage inference method similar to the re-rank mechanism. These approaches lead to RaSa and APTM achieving optimal performance, however, introducing additional training and inference costs. Instead, our PLIP achieves competitive performance without bells and whistles. Specifically, with ResNet50 as backbone, PLIP outperforms LGUR by 6.9\% and 7.5\% rank-1 on each dataset respectively.

\vspace{-5mm}
\subsection{Evaluation on Image-based Re-ID}
\vspace{-2mm}

\begin{wraptable}[13]{r}{7cm}
\vspace{-26mm}
\centering
\caption{Comparison on two baseline methods by using different pre-trained models. The best results are shown in bold.}
\label{tab:unsupreid}
\resizebox{\linewidth}{!}{
\begin{tabular}{c|c|cccc|cccc}
\hline
\multicolumn{1}{c|}{\multirow{2}{*}{}} & \multicolumn{1}{c|}{\multirow{2}{*}{Pre-train}}&\multicolumn{4}{c|}{Market1501}&\multicolumn{4}{c}{MSMT17}\\
\cline{3-10}
& & mAP & R@1& R@5& R@10 &mAP&R@1&R@5&R10 \\
\hline
\multicolumn{1}{c|}{\multirow{6}{*}{\rotatebox{90}{PPLR~\cite{PPLR}}}}
&{Baseline} &81.5&92.8 &97.1&98.1&31.4&61.1 &73.4&77.8 \\
&{MoCov2~\cite{mocov2}}&79.6&91.6&96.6&97.9&28.7&57.5&69.6&74.6\\
&{CLIP~\cite{CLIP}} &75.0&89.0 &95.6&97.1&7.6&20.3 &29.9&34.9\\
&{LUP~\cite{LUP}} &57.5&78.2 &85.9&88.9&22.5&48.9 &62.0&66.9\\
&{LUP-NL~\cite{LUPnl}} &84.2&93.4 &97.7&98.6&25.0&50.3 &63.1&68.7\\

&{\cellcolor{gray!20}PLIP}&\cellcolor{gray!20}\textbf{86.6}&\cellcolor{gray!20}\textbf{94.5} &\cellcolor{gray!20}\textbf{98.0}&\cellcolor{gray!20}\textbf{98.7}&\cellcolor{gray!20}\textbf{46.1}&\cellcolor{gray!20}\textbf{73.4} &\cellcolor{gray!20}\textbf{83.7}&\cellcolor{gray!20}\textbf{87.0}\\

\hline
\multicolumn{1}{c|}{\multirow{6}{*}{\rotatebox{90}{ISE~\cite{ISE}}}}
&{Baseline} &84.7&94.0 &97.8&98.8&35.0&64.7 &75.5&79.4\\
&{MoCov2~\cite{mocov2}}&84.9&93.5&97.6&98.7&34.1&64.5&75.0&79.0\\
&{CLIP~\cite{CLIP}} &79.5 &92.0&96.8&98.0&14.9 &35.2&46.3&51.3\\
&{LUP~\cite{LUP}} &84.5&94.2 &97.6&98.4&27.8&56.7 &68.8&73.4\\
&{LUP-NL~\cite{LUPnl}} &87.4&95.0 &98.2&98.9&33.9&62.5 &73.2&77.3\\

&{\cellcolor{gray!20}PLIP} &\cellcolor{gray!20}\textbf{87.6}&\cellcolor{gray!20}\textbf{95.3} &\cellcolor{gray!20}\textbf{98.2}&\cellcolor{gray!20}\textbf{98.9}&\cellcolor{gray!20}\textbf{46.4}&\cellcolor{gray!20}\textbf{74.9} &\cellcolor{gray!20}\textbf{84.3}&\cellcolor{gray!20}\textbf{87.5}\\
\hline
\end{tabular}}
\end{wraptable}

\textbf{Unsupervised methods achieve significant
improvements.} With simply replacing the backbone, our pre-trained models benefit unsupervised image-based person Re-ID methods significantly. We evaluate the improvement brought by different pre-trained ResNet50 models to the SoTA unsupervised methods PPLR~\cite{PPLR} and ISE~\cite{ISE}. As shown in Tab.~\ref{tab:unsupreid}, PLIP outperforms all other pre-trained models by a large margin. Specifically, applied to ISE, PLIP achieves new SoTA performance, outperforming the previsous SoTA by 2.9\% and 11.4\% mAP on Market1501 and MSMT17, respectively.

\begin{wraptable}[11]{r}{7cm}
\vspace{-30mm}
\centering
\caption{Comparison with SoTA methods on fully supervised image-based person Re-ID. The best results are shown in bold.}
\vspace{-1mm}
\label{tab6}
\resizebox{\linewidth}{!}{
\begin{tabular}{c|cc|cc|cc}
\hline
\multicolumn{1}{c|}{\multirow{2}{*}{Method}} & \multicolumn{2}{c|}{Settings}&\multicolumn{2}{c|}{Market1501}&\multicolumn{2}{c}{DukeMTMC}\\
\cline{2-7}
&Backbone&Pretrain & mAP & R@1 & mAP & R@1  \\
\hline

\multicolumn{1}{c|}{MGN~\cite{MGN}}&RN50&IMG &87.5 &95.1 &79.4&89.0 \\

\multicolumn{1}{c|}{ABDNet~\cite{ABDNet}}&RN50&IMG &88.3 &95.6 &78.6&89.0\\

\multicolumn{1}{c|}{GCP~\cite{GCP}}&RN50&IMG &88.9 &95.2 &78.6&87.9\\

\multicolumn{1}{c|}{ISP~\cite{ISP}}&RN50&IMG &88.6 &95.3 &80.0&89.6\\

\multicolumn{1}{c|}{TransReID~\cite{transreid}}&ViT-B &IMG&88.2&95.0 &80.6 &89.6\\

\hdashline[2.5pt/5pt]
\multicolumn{1}{c|}{UPReID~\cite{upreid}}&RN50 &LUP&91.1&97.1 &- &- \\

\multicolumn{1}{c|}{LUP~\cite{LUP}}&RN50 &LUP&91.0&96.4 &82.1 &91.0 \\

\multicolumn{1}{c|}{LUP-NL~\cite{LUPnl}}&RN50 &LUP-NL&91.9&96.6 &84.3 &92.0 \\

\multicolumn{1}{c|}{PASS~\cite{pass}}&ViT-B &LUP&93.0&96.8 &- &- \\

\hline
\multicolumn{1}{c|}{PLIP}&RN50&SYNTH &91.4 &96.8  &81.7 &91.1 \\
\multicolumn{1}{c|}{PLIP}&RN101&SYNTH & 92.0  &96.9  &82.3&91.8\\
\multicolumn{1}{c|}{PLIP}&RN152&SYNTH &92.6  &97.1  &83.1 &92.1 \\
 \rowcolor{gray!20}
\multicolumn{1}{c|}{PLIP}&Swin-B&SYNTH &\textbf{93.2}&\textbf{97.3}  &\textbf{84.4} &\textbf{92.2} \\
\hline
\end{tabular}}
\end{wraptable}
\textbf{Comparison with state-of-the-art methods.} 
We compare our results with existing SoTA image-based person Re-ID methods on Market1501 and DukeMTMC. Any results gained from post-processing techniques like re-rank~\cite{rerank} are excluded for a fair comparison. As indicated in Tab.~\ref{tab6}, by applying our PLIP on ABD-Net~\cite{ABDNet}, we achieve competitive performance on all datasets. Moreover, we achieve new SoTA performance with Swin-Base as the backbone. This demonstrates that the learned representations benefits this task significantly.

\vspace{-2mm}
\subsection{Evaluation on Person Search}
\vspace{-2mm}

\begin{wraptable}[13]{r}{7cm}
\vspace{-32mm}
\tabcolsep=3.2pt
\tiny
\caption{Comparison on two baseline methods by using different pre-trained models. We show the best score in bold.}
\vspace{-2mm}
\label{person_search_pretrain}
\resizebox{\linewidth}{!}{
\begin{tabular}{c|c|cccc|cccc}
\hline
\multicolumn{1}{c|}{\multirow{2}{*}{}} & \multicolumn{1}{c|}{\multirow{2}{*}{Pre-train}}&\multicolumn{4}{c|}{CUHK-SYSU}&\multicolumn{4}{c}{PRW}\\
\cline{3-10}
& & mAP & R@1& R@5& R@10 &mAP&R@1&R@5&R10 \\
\hline
\multicolumn{1}{c|}{\multirow{6}{*}{\rotatebox{90}{SeqNet~\cite{SeqNet}}}}
&{Baseline} &94.8 &95.7&98.1&98.7&47.6 &87.6&94.4&95.4 \\
&{MoCov2~\cite{mocov2}} &94.0&94.4 &98.2&98.5&48.3&87.3 &94.6&95.8\\
&{CLIP~\cite{CLIP}} &92.8&93.4 &97.8&98.2&48.0&86.7 &94.5&95.9\\
&{LUP~\cite{LUP}} &94.3&94.9 &98.2&98.7&47.7&87.3 &94.5&95.7\\
&{LUP-NL~\cite{LUPnl}} &95.4&95.9 &98.2&98.8&49.3&87.5 &94.3&95.9\\

&{\cellcolor{gray!20}PLIP}&\cellcolor{gray!20}\textbf{96.0}&\cellcolor{gray!20}\textbf{96.7} &\cellcolor{gray!20}\textbf{98.7}&\cellcolor{gray!20}\textbf{99.1}&\cellcolor{gray!20}\textbf{53.5}&\cellcolor{gray!20}\textbf{88.9} &\cellcolor{gray!20}\textbf{95.2}&\cellcolor{gray!20}\textbf{96.3}\\

\hline
\multicolumn{1}{c|}{\multirow{6}{*}{\rotatebox{90}{GLCNet~\cite{GLCNet}}}}
&{Baseline} &95.8&96.2 &98.3&98.8&47.8&87.8 &94.5&95.5\\
&{MoCov2~\cite{mocov2}} &95.1&95.6 &98.0&98.4&48.4&87.8 &94.6&95.9\\
&{CLIP~\cite{CLIP}} &93.0&93.6 &97.7&98.1&48.2&87.1 &94.5&96.0\\
&{LUP~\cite{LUP}} &95.5&95.8 &98.3&98.9&48.1&87.7 &94.5&95.8\\
&{LUP-NL~\cite{LUPnl}} &96.0&96.4 &98.4&99.0&49.8&87.8 &94.6&96.0\\

&{\cellcolor{gray!20}PLIP} &\cellcolor{gray!20}\textbf{96.3}&\cellcolor{gray!20}\textbf{97.0} &\cellcolor{gray!20}\textbf{99.0}&\cellcolor{gray!20}\textbf{99.3}&\cellcolor{gray!20}\textbf{53.7}&\cellcolor{gray!20}\textbf{89.0} &\cellcolor{gray!20}\textbf{95.4}&\cellcolor{gray!20}\textbf{96.4}\\

\hline
\end{tabular}}
\end{wraptable}

\textbf{Improvement over existing methods.} 
Our pre-trained models bring significant performance gain to existing person search methods. To verify this, we choose two representative methods SeqNet~\cite{SeqNet} and GLCNet~\cite{GLCNet} as the baselines, and evaluate the improvements brought by different pre-trained models. As shown in Tab.~\ref{person_search_pretrain}, under the ResNet50 setting, our model brings maximum performance improvements to all methods. Specifically, when applying our pre-trained ResNet50 to GLCNet, it achieves 96.3\% and 53.7\% mAP on the CUHK-SYSU and PRW datasets, respectively.

\begin{wraptable}[13]{r}{7cm}
\vspace{-36.5mm}
\tiny
\centering
\caption{Comparison with state-of-the-art methods on the end-to-end person search. We show the best score in bold.}
\vspace{-2mm}
\label{person_search_sota}
\resizebox{\linewidth}{!}{
\begin{tabular}{c|c|cc|cc}
\hline
\multicolumn{1}{c|}{\multirow{2}{*}{Method}} & \multicolumn{1}{c|}{\multirow{2}{*}{Backbone}}&\multicolumn{2}{c|}{CUHK-SYSU}&\multicolumn{2}{c}{PRW}\\
\cline{3-6}
& & mAP & R@1 & mAP & R@1  \\
\hline
\multicolumn{1}{c|}{OIM~\cite{OIM}}&RN50 &75.5 &78.7 &21.3&49.9 \\

\multicolumn{1}{c|}{NPSM~\cite{NPSM}}&RN50 &77.9 &81.2 &24.2&53.1 \\ 

\multicolumn{1}{c|}{CTXGraph~\cite{CTXGRAPH}}&RN50 &84.1 &86.5 &33.4&73.6\\ 

\multicolumn{1}{c|}{QEEPS~\cite{QEEPS}}&RN50 &88.9 &89.1 &37.1&76.7\\

\multicolumn{1}{c|}{BINet~\cite{BINet}}&RN50 &90.0 &90.7 &45.3&81.7\\

\multicolumn{1}{c|}{NAE+~\cite{NAE}}&RN50 &92.1 &92.9 &44.0&81.1\\

\multicolumn{1}{c|}{SeqNet~\cite{SeqNet}}&RN50 &94.8 &95.7 &47.6&87.6\\

\multicolumn{1}{c|}{GLCNet~\cite{GLCNet}}&RN50 &95.8 &96.2 &47.8&87.8\\

\multicolumn{1}{c|}{SOLIDER~\cite{SOLIDER}}&Swin-B &94.9 &95.5 &\textbf{59.7}&86.8\\
\hline
\multicolumn{1}{c|}{PLIP}&RN50 & 96.3  &97.0  &53.7	&89.0  \\
\multicolumn{1}{c|}{PLIP}&RN101 & 96.5  & 97.2 &55.1	&89.4  \\
\multicolumn{1}{c|}{PLIP}&RN152 &\textbf{96.7}  &97.3  &56.2	&89.9  \\
 \rowcolor{gray!20}
\multicolumn{1}{c|}{PLIP}&Swin-B & 96.6  &\textbf{97.5}  &57.8	&\textbf{90.3} \\
\hline
\end{tabular}}
\end{wraptable}
\textbf{Comparison with state-of-the-art methods.} 
We compare our results with existing SoTA person search methods in Tab.~\ref{person_search_sota}. By applying our pre-trained models on GLCNet~\cite{GLCNet}, we achieve new SoTA performance on each dataset. Specifically, under ResNet50 setting, our method surpasses the previous SoTA GLCNet by 5.9\% mAP on PRW dataset. Also, with Swin-Base as the backbone, we achieve the best performance compared with all other methods. These results demonstrate that our pre-training framework shows great potential in learning discriminative person representation for this task.

\vspace{-2mm}
\subsection{Ablation studies and analyses}
\vspace{-2mm}
\begin{wraptable}[9]{r}{7cm}
\vspace{-41mm}
\centering
\caption{Ablation study on the effectivenss of each pretext task, all using default settings.}
\vspace{-2mm}
\label{tab9}
\resizebox{\linewidth}{!}{
\begin{tabular}{c|ccc|ccc|ccc}
\hline 
\multicolumn{1}{c|}{\multirow{2}{*}{\#}} & \multicolumn{3}{c|}{Components}&\multicolumn{3}{c|}{CUHK-PEDES}&\multicolumn{3}{c}{Market1501}\\
\cline{2-10}
& IVLC &TIC &IAP&R@1&R@5&R@10&R@1&R@5&R@10    \\
\hline 

1 &\checkmark& &&30.2	&53.3&64.0	&62.3&79.9	&85.7 \\
2 &\checkmark& \checkmark& &31.4	&55.2&65.9	&62.9&80.7	&86.2 \\
3&\checkmark& &\checkmark&30.5	&54.4&64.3	&62.7&80.5	&86.1 \\
4&&\checkmark&\checkmark&-&-&-&39.3	&61.4&70.4	 \\
\rowcolor{gray!20}
5&\checkmark&\checkmark&\checkmark&\textbf{32.5}	&\textbf{56.3}&\textbf{66.6}	&\textbf{63.1}&\textbf{80.8}	&\textbf{86.3} \\
\hline 
\end{tabular}}
\end{wraptable}
We perform ablation studies with ResNet-50 as the visual encoder and pre-training on the sub-dataset of SYNTH-PEDES, which has 10,000 identities, 139,564 images and 353,617 textual descriptions. To assess the \textbf{effectiveness of pretext tasks} on the generalizability of our models, we directly evaluate the zero-shot performance of pre-trained models on downstream datasets. As we can see in Tab.~\ref{tab9}, each task contributes to the model's zero-shot capability and combining all of the tasks leads to the best performance. 

\begin{wrapfigure}[0]{r}{7cm}
\vspace{-52mm}
\centering
	\includegraphics[width=\linewidth]{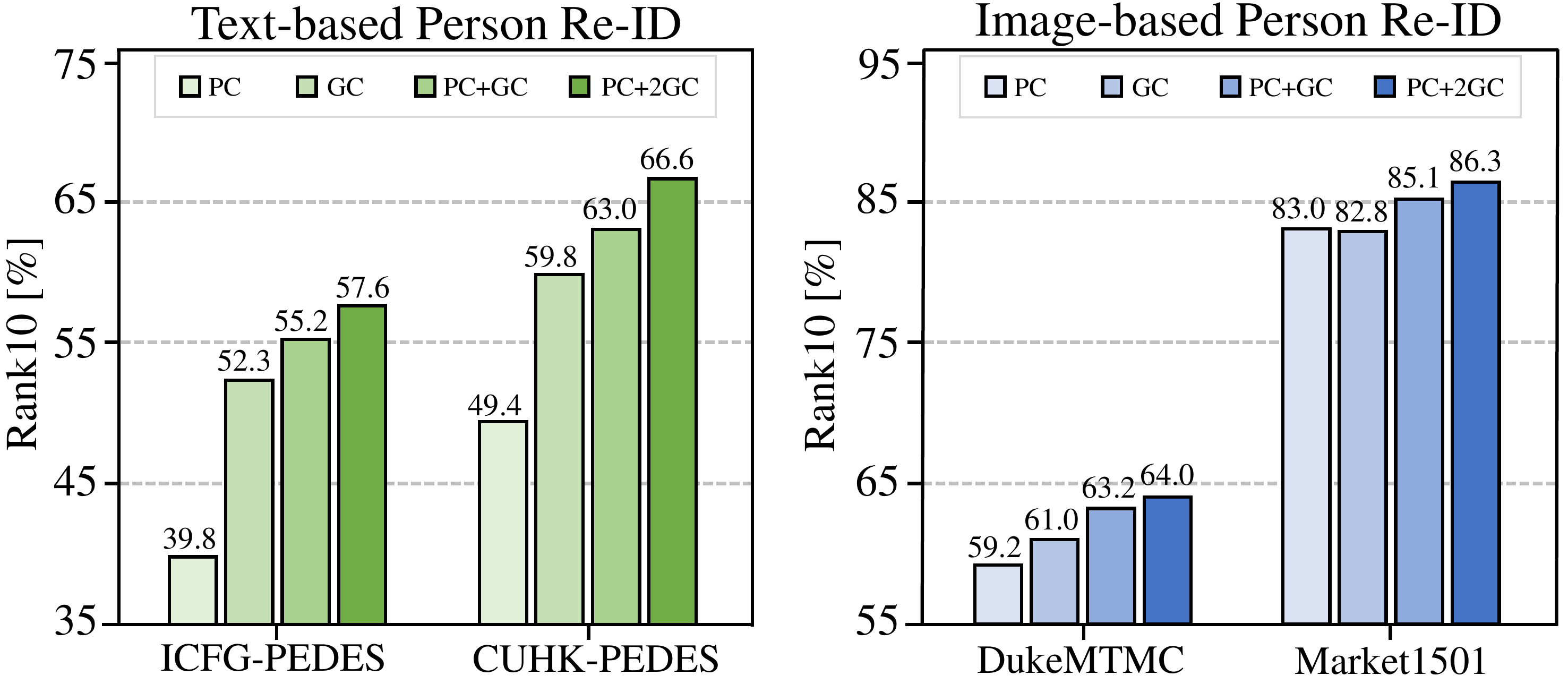}
 \vspace{-6mm}
\caption{The diversity of textual descriptions matters. PC and GC mean prompt caption and generated caption, respectively.}
\vspace{-5mm}
\label{diversity_matters}
\end{wrapfigure}
\vspace{-2mm}
Meanwhile, to validate the \textbf{effectiveness of textual diversity} in SYNTH-PEDES, we have studied four different degrees of textual diversity from weak to strong. As shown in Fig.~\ref{diversity_matters}, the fourth case with the highest degree of textual diversity has the best performance. More thorough ablation studies and analyses about effectiveness of each component, dataset quality evaluation, pre-training settings and so on can be found in Sec.~\ref{throughablation} of the appendix.

\vspace{-3mm}
\section{Conclusion}
\vspace{-3mm}
In this paper, we propose a novel language-image self-supervised person representation learning framework named PLIP, which consists of three well-motivated pretext tasks. Also, we build a large-scale real-scenario image-text person dataset SYNTH-PEDES by auto-captioning procedures. We achieve good generic person representation learning by utilizing PLIP and SYNTH-PEDES. Equipped with our pre-trained models, we push many existing methods to a much higher level without bells and whistles. We hope that our simple and effective framework can inspire researchers to devote further attention to this area.

\newpage
\bibliographystyle{plain}
\bibliography{main}

\newpage
\appendix
\section{Appendix}

\subsection{Related Work}
\textbf{General Representation Learning.}
To avoid the labor-intensive manual annotation process, many general representation learning approaches have been explored to utilize unlabeled image data. Several representative works~\cite{mocov1,simmim,simclr,byol,simsiam,beit,mae,dino,CAE,WJD2} have achieved performance comparable to, or even surpassing, supervised methods. For example, CAE~\cite{CAE} presents a novel masked image modeling approach and shows the benefit to representation learning through an encoder-regressor-decoder architecture. Besides to pre-training on unlabeled image data, there is currently a popular trend of attempting to establish the relationship between vision and language to learn more discriminative representations. For example, CLIP~\cite{CLIP} and ALIGN~\cite{Align} perform cross-modal contrastive learning on hundreds or thousands of millions of image-text pairs crawled from the web and can directly perform open-vocabulary image classification. BLIP~\cite{BLIP} shows that the noisy web texts are suboptimal for vision-language learning, and proposes a Captioning and Filtering method to improve the quality of the text corpus.

\textbf{Person Representation Learning.}
In person related field, it is common practice to leverage the backbones pre-trained on ImageNet~\cite{imagenet}, which ignores domain gap between general images and person-related images, and leads to limited performance. To address such a problem, LUP~\cite{LUP} constructs a large-scale unlabeled person dataset and makes the first attempt of performing a general unsupervised pre-training method MoCov2~\cite{mocov2} for learning person representations. The good experimental results on the person Re-ID task verified the effectiveness of it. 

However, simply adopting the general pre-training method may result in the ignorance of person fine-grained characteristics. Therefore, considering the particularity of Re-ID tasks, the proposed UP-ReID~\cite{upreid} introduces an intra-identity regularization and is the first attempt toward a Re-ID specific pre-training framework by explicitly pinpointing the difference between the general pre-training and Re-ID pre-training. Before long, LUP-NL~\cite{LUPnl} develop a large-scale pre-training framework utilizing noisy labels, and demonstrate that learning directly from raw videos is a promising alternative for pre-training, which utilizes spatial and temporal correlations as weak supervision. This simple pre-training task provides a scalable way to learn good Re-ID representations from scratch without bells and whistles. At the same time, PASS~\cite{pass} is proposed and it is more suitable for Re-ID by using several learnable tokens to extract the part-level features offering fine-grained information. It helps the ViTs set the new state-of-the-art performance on several person Re-ID datasets.

However, these approaches are only target at learning person representations especially for promoting the person Re-ID performance. They have poor generalization ability and perform poorly on other person-centric tasks. To learn a general human representation, the SOLIDER~\cite{SOLIDER} is proposed recently. It takes advantages of prior knowledge to produce pseudo semantic labels, and utilize them to train the representation with more semantic information through a semantic controller. 

In general, these pure-vision based works are rather aimed at only promoting the person Re-ID performance, or limited to the person-centric visual tasks. Their performance in learning generic person representation is still unsatisfactory and lacks the ability of multi-modal understanding.

\textbf{Person-centric Tasks.}
In computer vision community, there are many tasks directly or indirectly related to person, \ie image/text based person Re-ID, person attribute recognition, person pose estimation, person search, multi-object tracking and human parsing. We name such tasks as person-centric tasks and sort out five representative tasks among them for study.

1) \textit{Text-based person Re-ID} aims to search for person images of a specific identity by textual descriptions. Existing works can be divided into attention-based and attention-free methods. The former~\cite{SMAT,textreid,MG,SSAN, LGUR} attempts to establish region-text correspondences but ignores the efficiency. To better align the multi-modal features, the latter usually focuses on designing various objective functions~\cite{CMPM,vse++,ARL} and model structures~\cite{vitaa,dualpath}. 

2) \textit{Image-based person Re-ID} aims to search for person images by given person images. Most works are based on supervised learning. The hard triplet loss~\cite{triplet1,triplet2,triplet3} and classification loss~\cite{SGGNN,DCNN} are introduced to learn a global feature. Also, some works~\cite{PABR,BPM} focus on learning a part-based feature instead. For example, Sun $\textit{et al.}$~\cite{BPM} proposed to represent features as horizontal stripes and learn with separate classification losses. Some works utilize camera style~\cite{CID} and viewpoints~\cite{VTM} to train a more robust model. Some works~\cite{shi2023dual,shi2024multi} focus on semi-supervised or unsupervised setting.

3) \textit{Person attribute recognition} aims to identify the person’s attributes. Many methods~\cite{a1,a2,a3} treat this task as a multi-label classification problem, while some others~\cite{a4,a5,a6} adopt recurrent neural networks for exploring the attribute context. Also, some works~\cite{VTB,Label2Label} introduce an additional language modality to get better performance. 

4) \textit{Person search} aims to jointly localize and identify a query person from natural and uncropped images. Existing works can be generally categrized into two-step and one-step ones. In two-step works~\cite{igpn,tcts}, persons are detected and then fed into a Re-ID model for identification, while one-step approaches~\cite{SeqNet,GLCNet,NormNet} makes the joint framework more effective and efficient. 

5) \textit{Human parsing} is a fine-grained semantic segmentation task in human analysis. Some early works~\cite{LIP,jointmulti} combine human parsing with pose estimation. Moreover, some works~\cite{multihuman,graphonomy} study the human parsing task in a multi-person scenario, which needs to distinguish the human instances. Meanwhile, the work~\cite{SCHP} proposes a self-correction mechanism and leads to better performance.

\subsection{Broader Impact}
\label{broad_impact}
This paper proposes a language-image pre-training framework for person representation learning, and contributes a large-scale synthetic person-centric dataset with rich image-text pairs. Our pre-trained models and dataset can help diverse person-related applications such as person re-identification, person attribute recognition and human parsing, thus boosting the development of smart retail, smart transportation, smart security systems and so on in the future metropolises. 

Nevertheless, the application of person Re-ID and attribute recognition, such as for recognizing and tracking pedestrians in surveillance systems, might raise privacy concerns. It typically depends on the utilization of surveillance data for training without the explicit consent of the individuals being recorded. Therefore, governments and officials need to carefully establish strict regulations and laws to control the usage of these technologies. Otherwise, these technologies can potentially equip malicious actors with the ability to surveil pedestrians through multiple cameras without their consent. Furthermore, we should be cautious of the misidentification of the Re-ID systems to avoid possible disturbance. Also, note that the demographic makeup of the datasets used is not representative of the broader population.

At the same time, we have utilized a substantial amount of person-containing video data from the internet for pre-training purposes. Consequently, it is inevitable that the resulting models may inherently contain information about these persons. Researchers should adhere to relevant laws and regulations, and strive to avoid using our models for any improper invasion of privacy. We will have a gated release of our models and training data to avoid any misuse. We will require that users adhere to usage guidelines and restrictions to access our models and training data. Meanwhile, all our open-sourced assets can only be used for research purpose and are forbidden for any commercial use.

\subsection{Limitations}
\label{limitations}
PLIP presents a preliminary attempt to introduce language modality into generic person representation learning. Despite its effectiveness on existing public datasets, PLIP may still be difficult to learn good fine-grained person representations for it does not explicitly achieve local information correlation across different modalities. Also, its pretext tasks require multiple forward propagation, directly increasing the memory overhead and affecting computational efficiency. Meanwhile, as we have followed the conventional practice of previous works by utilizing a large amount of person-containing internet video data to implement pre-training, there is a potential for privacy and security issues to some extent. Therefore, in our subsequent work, we will focus on addressing fine-grained issues and improving the efficiency of our framework. We will take every possible measure to prevent the misuse of our models and dataset as well.

\subsection{Altering Color Word Affects Image Colorization: Visualization}
\begin{figure}[h]
\centering
\includegraphics[width=\linewidth]{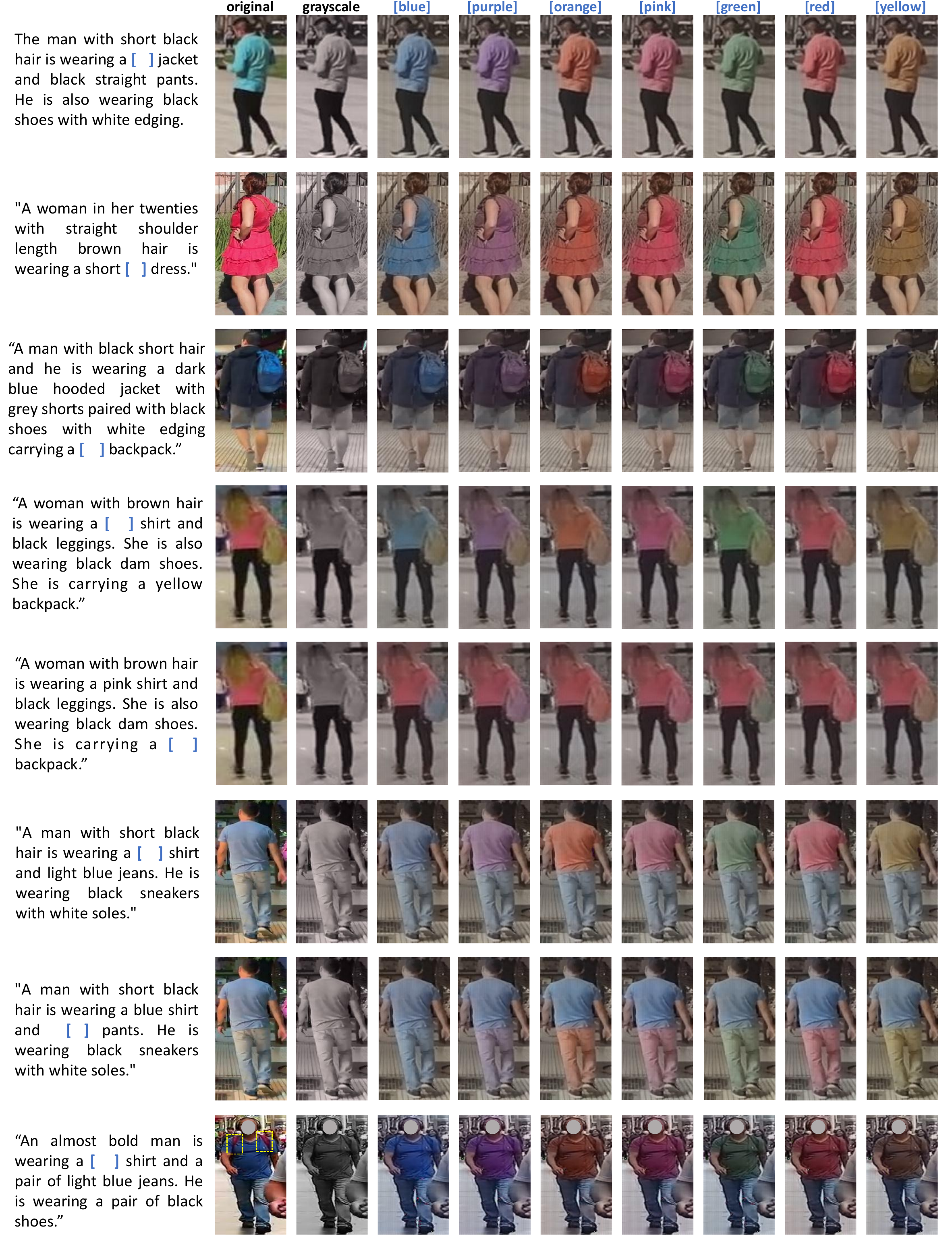}
\caption{Visualization of gray-scale person image colorization results by changing the color words in textual descriptions.}
\label{fig:sic_colorize}
\end{figure}

As displayed in Fig~\ref{fig:sic_colorize}, in our pretext task of text-guided image colorization, altering the color word in textual description significantly affects the colorization of image. 
However, our model may not fully understand the semantics of more detailed image regions. As shown in the last row, our model fails to distinguish between the blue clothing region and the red shoulder strap region (marked with yellow boxes), instead blending the two into a unified coloration.
This is due to the fact that the level of detail in manually annotated datasets is still not sufficient, resulting in the model being unable to theoretically learn representations with higher levels of detail and greater discrimination capabilities.
However, it is undeniable that our model has a preliminary understanding of the meaning of attributes and colors, and can associate them with related image regions, rather than simple memorization. This ability to distinguish between different parts of the person body guarantees the superior performance on the subsequent person-centric tasks. 

\newpage
\subsection{How to Construct SYNTH-PEDES and its Properties}
\label{howtoconstructsynthpedes}
We build the SYNTH-PEDES dataset to pre-train the PLIP models at a large-scale. In this section, we show the details of how to construct our SYNTH-PEDES dataset and its characteristic properties.

\subsubsection{Dataset Construction}
\label{datasetconstructdetail}
The dataset construction process can be described as three steps. Firstly, we collect several person datasets~\cite{LUPnl,lpw} to form the large-scale image dataset. Secondly, as a person image captioner, a simple but effective method is proposed to achieve automatic diversified text annotation with high-quality. Finally, we adopt some post-processing approaches to eliminate the noises and improve the dataset quality.

\textbf{Image Collection.} 

We collect and process two large-scale person datasets to form the image dataset. The first is LUPerson-NL~\cite{LUPnl}. It is a new variant of LUPerson~\cite{LUP} on top of raw videos from LUPerson and assign the noisy labels to each person image with automatically generated tracklet. It consists of 10\textit{M} images with about 430\textit{K} identities collected from 21\textit{K} scenes. The second is LPW~\cite{lpw}. It consists of 2,731 different persons and 592,438 images collected from three different crowded scenes. 

\begin{wrapfigure}[17]{r}{7cm}
\vspace{-5mm}
\centering
	\includegraphics[width=\linewidth]{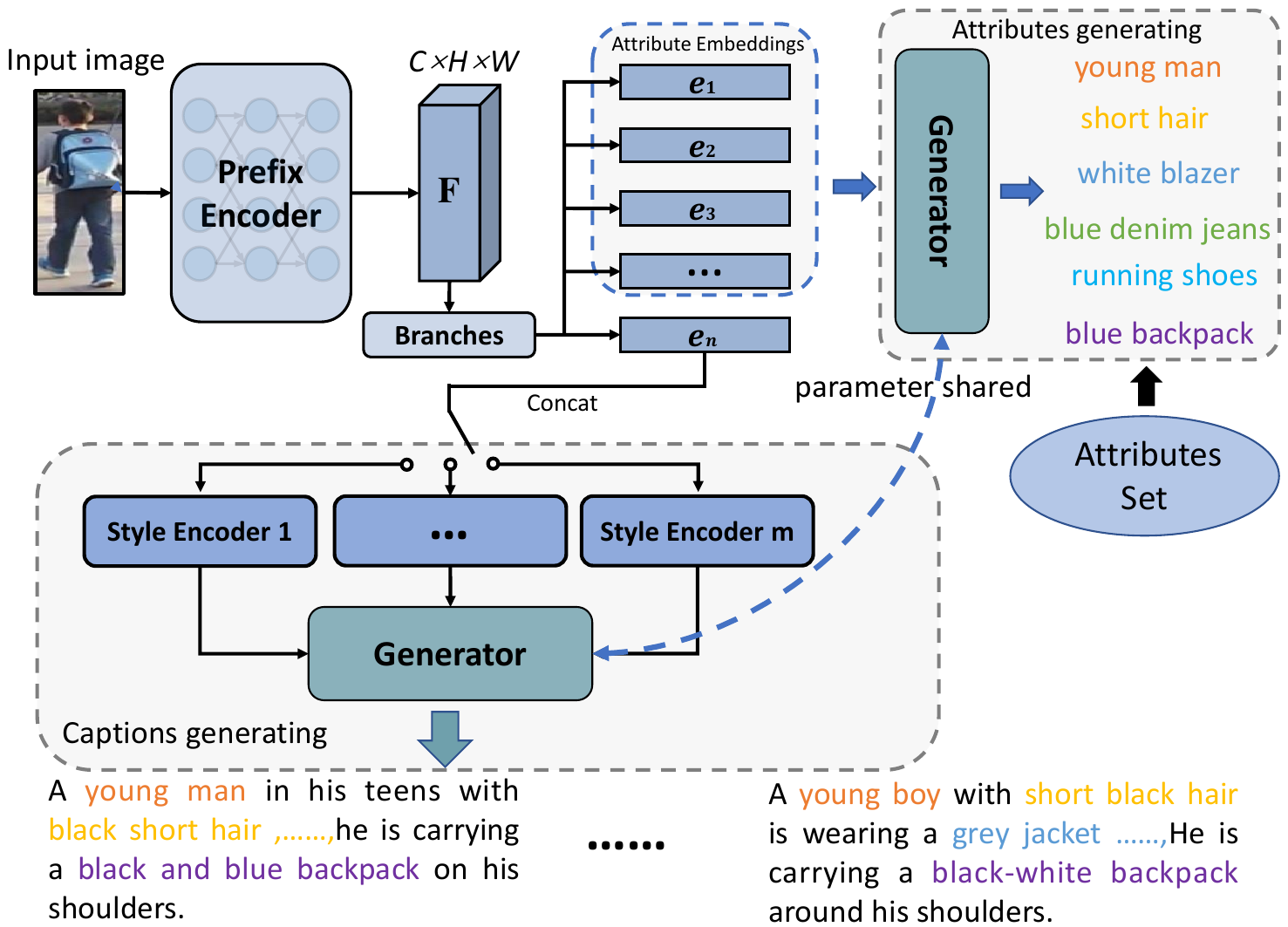}
\caption{Architecture of Stylish Pedestrian Attributes-union Captioning. It mainly consists of a prefix encoder and a generator. We use it to generate descriptions from a person image.}
\label{fig:SPAC}
\end{wrapfigure}

\textbf{Image Captioner (SPAC).} 

Given an input person image, there is no specific work targeting at generating captions that detailedly describe the person's appearance. To this end, we propose a simple but effective method for person image captioning. It can generate attribute-annotations and stylish textual-descriptions, which simulate the diverse perspectives that different annotators may have on the same person image. 

As illustrated in Fig.~\ref{fig:SPAC}, Stylish Pedestrian Attributes-union Captioning (SPAC) mainly comprises two modules, \ie a prefix encoder and a shared generator. Specifically, we use ResNet101-FPN~\cite{resnet,fpn} as the encoder and GPT2~\cite{gpt2} as the generator to capture rich person image details and generate high-quality texts.
In the existing image-text person datasets~\cite{textreid,SSAN}, many unique workers were involved in the labeling tasks. The same images usually have inconsistent-style language descriptions. Constructing image-text pairs for an image with multiple descriptions will lead to unstable training and affect the model performance. Thus, in order to replicate the stylized variations that arise from multiple workers' labels and mitigate the issue of redundant real labels, we have incorporated style encoders into our pipeline. We let the concatenated prefixes pass through different style encoders to get the prefixes of their own style and then send them to the generator for the subsequent generation.

The entire training process can be seen as an autoregressive problem. Given a dataset of paired images, attributes and captions $\left\{\mathbf{x}^{i} , \mathbf{A}^{i}, \mathbf{y}^{i}\right\}_{i=1}^{N}$, where $\mathbf{A}^{i}$ is an attribute set of image $\mathbf{x}^{i}$ containing six attribute descriptions, the learning goal is to generate meaningful attribute descriptions and captions from an unseen person image. The attributes and captions can be referred as a sequence of padded tokens $\mathbf{A}^i=\left\{\mathbf{a}_1^{i,k},\ldots,\mathbf{a}_{\ell_1}^{i,k}\right\}_{k=1}^{n-1},\mathbf{y}^i={\mathbf{y}_1}^i,\ldots,{\mathbf{y}_{\ell_2}}^i$, with maximal lengths $\ell_1,\ell_2$ accordingly. 

Following recent works~\cite{clipcap,zhou2019vlp}, our key solution is to jointly train a prefix encoder and a generator. The former is to capture the semantic embeddings as the prefixes from the image, and the latter, as an autoregressive language model, is to use the prefixes to predict the next token one by one. As shown in Fig.~\ref{fig:SPAC}, we first feed the input image $\mathbf{x}^i$ into the prefix encoder $PE$ and different branches $\left\{BR_k\right\}_{k=1}^{n}$ to get the $n$ attribute-and-relation prefix embeddings:

\begin{equation}\label{eq1}
    \mathbf{e}_1^i,\ldots,\mathbf{e}_n^i=\left\{BR_k(PE(\mathbf{x}^i))\right\}_{k=1}^{n}.
\end{equation}

And then the concatenated embeddings are fed into different style encoders $\left\{SE_k\right\}_{k=1}^m$ to get the stylized caption embeddings:
\begin{equation}\label{eq2}
    \mathbf{c}_1^i,\ldots,\mathbf{c}_m^i=\left\{SE_k(concat([\mathbf{e}_1^i,\ldots,\mathbf{e}_n^i]))\right\}_{k=1}^{m},
\end{equation}
each embedding has the same dimension as a token embedding. We then concatenate the obtained embeddings to the atrribute and caption token embeddings, where $\mathbf{e}^i$ is selected from the stylized caption embeddings in turn:
\begin{equation}\label{eq3}
\begin{aligned}
    \{\mathbf{Z}_k^i &=concat([\mathbf{e}_k^i,\mathbf{a}_1^{i,k},\ldots,\mathbf{a}_{\ell_1}^{i,k}])\}_{k=1}^{n-1}, \\
    \mathbf{Z}_n^i &= concat([\mathbf{e}^i,{\mathbf{y}_1}^i,\ldots,{\mathbf{y}_{\ell_2}}^i].
\end{aligned}
\end{equation}

Finally, we feed the embeddings $\{\mathbf{Z}^i\}_{i=1}^N$ into the shared generator $G$ to predict the attribute and caption tokens in an autoregressive fashion, using the cross-entropy loss:

\begin{equation}\label{eq4}
    \begin{aligned}
    \mathcal{L}_{c}&=\sum_{i=1}^{N}\sum_{j=1}^{\ell_2} \log G\left(\mathbf{y}_{j}^{i} \mid \mathbf{e}^i, \mathbf{y}_{1}^{i}, \ldots, \mathbf{y}_{j-1}^{i}\right),
    \end{aligned}
\end{equation}

\begin{equation}\label{eq5}
    \begin{aligned}
\mathcal{L}_{a}&=\{\sum_{i=1}^{N}\sum_{j=1}^{\ell_1} \log G\left(\mathbf{a}_{j}^{i,k} \mid \mathbf{e}_k^{i}, \mathbf{a}_{1}^{i,k}, \ldots, \mathbf{a}_{j-1}^{i,k}\right)\}_{k=1}^{n-1}.
    \end{aligned}
\end{equation}

Define $\lambda \in \mathbb{R}^{+}$ as a balance factor, then the overall loss $\mathcal{L}_{spac}$ is computed as:
\begin{equation}\label{eq6}
    \begin{aligned}
   \mathcal{L}_{spac} = -\mathcal{L}_{c} - \lambda \mathcal{L}_{a}.
    \end{aligned}
\end{equation}

\vspace{-3mm}
\textbf{Post-Processing}.

\begin{wrapfigure}[20]{r}{7cm}
\vspace{-5mm}
\centering 
	\includegraphics[width=\linewidth]{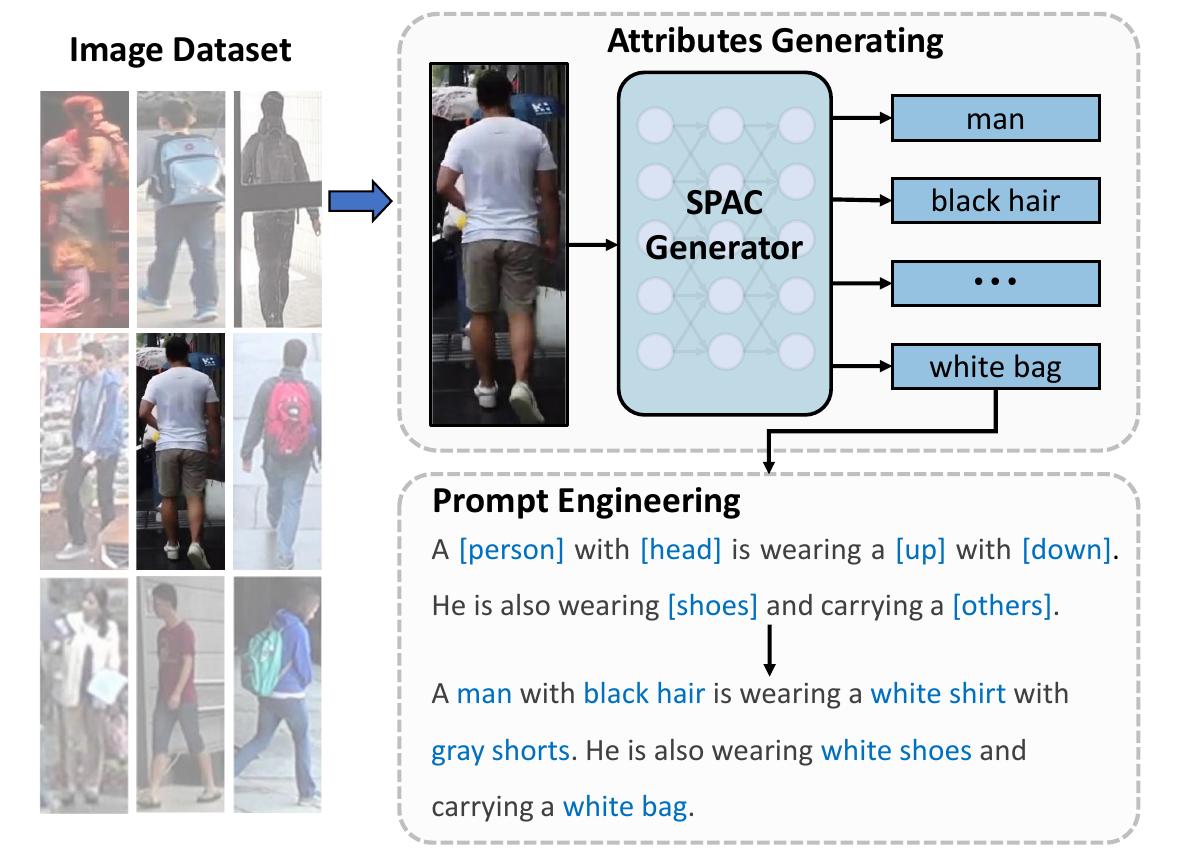}
\caption{Visual example of Attributes Prompt Engineering. Given an image, we generate the attribute annotations based on SPAC and embed them into the masked sentence randomly chosen from the sentence library to form a complete caption.}
\label{fig:Attributes_eng}
\vspace{-5mm}
\end{wrapfigure}
\textbf{Noise Filter Strategy.} To filter the noises in LUPerson-NL, we propose Seed Filter Strategy. Specifically, it mainly includes three processes. 1) \textit{Filter-out}. For all samples with the same identity, we cycle to calculate the similarity between the current sample and the center of other samples and exclude the samples whose similarity does not meet the threshold until the similarity of all samples with the same identity meets it. 2) \textit{Reassignment}. According to the similarity and threshold, we reassign the excluded samples in excluded dataset to the correct identity within a certain identity continual range. 3) \textit{Merger}. We merge the samples that should belong to the same identity but are divided into different identities. Through this process, the samples with incorrect identity label annotations can be well filtered out or included in their expected identity set. More details can be found in Sec.~\ref{Strategydetail}

\noindent
\textbf{Data Distribution Strategy.} There are some samples with poor imaging conditions in the image part. Also, the number of images per identity in LUPerson-NL is very unbalanced. To this end, we have adopted some strategies to ensure the quality of generated attributes, the consistency of gender annotation, and identity distribution balance. For example, we adopt a gender voting mechanism to automatically synchronize the gender annotation of an identity in dispute. For the text part, we aim to generate three sentences for an image. Two of them are directly generated by SPAC while the another is generated by Attributes Prompt Engineering exploiting the attribute annotations generated by SPAC, as shown in Fig.~\ref{fig:Attributes_eng}. There are averagely 2.53 sentence per image. More details can be found in Sec.~\ref{Strategydetail}

\vspace{-2mm}
\subsubsection{Dataset Properties}
\vspace{-1mm}
\begin{table*}[htb]
\centering
\caption{Statistics comparison on existing popular datasets. SYNTH-PEDES is by far the largest person dataset with textual descriptions without any human annotation effort.}
\label{table1}
\resizebox{\linewidth}{!}{
\begin{tabular}{c|c|c|c|c|c|c|c|c}
\hline 
Datasets & year &\#images &\#identities&\#descriptions&view &label type & label method  &crop size   \\
\hline 
Market1501~\cite{market} &2015& 32,668& 1,501&-	&fix camara &identity &hand+DPM~\cite{DPM}&128×64\\
DukeMTMC~\cite{dukemtmc}&2017&36,411& 1,852&-&fix camera &identity&hand&vary\\
CUHK-PEDES~\cite{textreid}&2017&40,206&13,003&80,412&fix camera &identity+description&hand&vary \\
LPW~\cite{lpw}&2018&592,438&2,731&-&fix camera &identity &hand+Detector+NN& vary\\
MSMT17~\cite{MSMT17}&2018&126,441&4,101&-&fix camera&identity&FasterRCNN~\cite{rcnn}&vary\\
SYSU30K~\cite{SYSU30K}&2020&29,606,918&30,508&-&dymamic&identity&YOLOv2~\cite{yolov2}&vary\\
RSTPReid~\cite{RSTPReid}&2021&20,505&4,101&41,010&fix camera&identity+description&hand&vary\\
ICFG-PEDES~\cite{SSAN}&2021&54,522&4,102&54,522&fix camera&identity+description&hand&vary\\
LUPerson~\cite{LUP}&2021&4,180,243&$> 200k$&-&dymamic&no&YOLOv5&vary\\
LUPerson-NL~\cite{LUPnl}&2022&10,683,716&433,997&-&dymamic&identity&FairMOT~\cite{fairmot}&vary\\
MALS~\cite{APTM}&2023&1,510,330&-&1,510,330&dynamic&description&ImaginAIry&vary\\
UFine6926~\cite{ufinebench}&2024&26,206&6,926&52,412&dynamic&identity+description&hand&vary\\
\hline
\cellcolor{gray!20}\textbf{SYNTH-PEDES}&\cellcolor{gray!20}2024&\cellcolor{gray!20}4,791,771&\cellcolor{gray!20}312,321&\cellcolor{gray!20}12,138,157&\cellcolor{gray!20}dymamic&\cellcolor{gray!20}identity+description&\cellcolor{gray!20}SPAC&\cellcolor{gray!20}vary\\
\hline 
\end{tabular}}
\end{table*}
Thanks to the outstanding generating ability of our proposed SPAC, the SYNTH-PEDES dataset is full of high-quality textual descriptions in a variety of styles, which can be utilized to train the representation learning model. Compared with existing person datasets in Tab.~\ref{table1}, SYNTH-PEDES has the following advantages: 

\noindent
\textbf{Diversified.} Our dataset contains a wide range of variations in the textual descriptions. Unlike the previous person datasets with only one or two image-text pairs, most images of our dataset are annotated with three textual descriptions. 

\noindent
\textbf{High-quality.} As some typical qualitative examples can be seen in Fig.~\ref{fig:captionsamples}, the generated annotations achieve an accurate and detailed description of the person appearance. The further experiments conducted on the quality research can be found in the following sections. Researchers can use this dataset with confidence to conduct relevant studies. 

\noindent
\textbf{Large-scale.} In Tab.~\ref{table1}, we compare the properties of SYNTH-PEDES with other popular person datasets. As we can see, SYNTH-PEDES is the largest real person dataset with high-quality image-text pairs by far, which contains 312,321 identities, 4,791,711 images, and 12,138,157 textual descriptions.

\subsection{Details of Network Structures}
In this section, we will provide a detailed explanation of the structure and some parameter settings of our representation learning network, mainly including some modifications made to the ResNet networks~\cite{resnet} and the decoder designs for the two pre-text tasks. The dimension of global embeddings is set to 768.
\subsubsection{Modifications to ResNets}
We made some modifications to the ResNet networks like CLIP~\cite{CLIP}. Firstly, there are now 3 "stem" convolutions as opposed to 1, with an average pool instead of a max pool. For each "stem" convolutions, it is consisted of a convolution layer, a batch normalization layer and a ReLU activation funtion. Secondly, we perform anti-aliasing strided convolutions, where an avgpool is prepended to convolutions with stride $>$ 1. Thirdly, the final pooling layer is a max pool instead of an average pool and we add a trainable linear layer to compress feature dimensions to 768.

\subsubsection{Specifications of TIC Decoder}
For an image with $3\times{H}\times{W}$, we firstly obtain the multi-scale visual feature map with $1024\times{\frac{H}{4}}\times{\frac{W}{4}}$ like FPN. Then, we send the feature map and textual global embedding to 2 Multimodal SE-Block and deconvolution layers to acquire the final feature map with $3\times{H}\times{W}$, which is utilized to restore the color information. Next, we specify the detailed designs of the Multimodal SE-Block.

In each MSE-Block, for a visual feature map with $C\times{H}\times{W}$, we firstly perform an average pooling operation on it to obtain a visual feature embedding with $1\times{C}$. Then we concat the visual feature embedding and the input texutal global embedding with $1\times{768}$ to obtain a fused feature embedding with $1\times{(C+768)}$. Then, we feed it to a FC layer, which contains a linear layer$((C+768),C)$, a ReLU activation, a linear layer$(C,C)$ and a sigmoid funtion, to obtain the fused attention embedding with $1\times{C}$. Finally, we apply it to the channel of the original feature map to obtain the multi-modal fused feature with $C\times{H}\times{W}$.

After each MSE-Block, there is a deconvolution layer operated on the output feature map. For the first block, the deconv is with input channel=1024, output channel=256, kernel size=3, stride=2, padding=1 and output padding=1. For the second block, the deconv is with input channel=256 output channel=3, and others the same.

After the approaches above, we finally obtain an output feature map with the same shape like the input original color image, and it can be used to predict the ground truth.

\subsubsection{Specifications of IAP Decoder}
For each masked token, we obtain its hidden state output as the masked embedding. We then simply concat the masked embedding and the visual global embedding to obtain the fused embedding. Then the fused embedding will be regarded as the QKV and set to a normal transformer block to improve the multimodal fusion. Finally, the output fused embedding of transformer block will be sent to a prediction head to predict the masked word. The prediction head is a simple linear layer with $(768, nvocab)$, where the $nvocab$ means the size of the dictionary.

\subsection{Details of Dataset Construction}
\label{Strategydetail}
In this section, we specify the details of our gathering SYNTH-PEDES dataset. We mainly provide a detailed explanation of the designs of our texutal description generated model SPAC. Also, we demonstrate the noise filter strategy and data distribution strategy adopted in the process of dataset constrution.
\subsubsection{Designs of SPAC Model}
The prefix encoder is a ResNet101-FPN~\cite{fpn} pre-trained on ImageNet~\cite{imagenet}. There are two types of prefix branches, one is the attribute branch and the other is the sentence branch. The detailed structures of each branch are shown in Tab.~\ref{branch_structure}. Prefix dimension is set to 768. Attribute prefix length and sentence prefix length are 3 and 5, respectively. 

There are totally 6 types of attribute prefix. For each type of attribute prefix, we send it to the generator GPT2~\cite{gpt2} and generate the attribute annotation accordingly. Also, we concat all the attribute prefixes and sentence prefix, and send them to 2 types of style encoders for obtaining the stylized prefixes. Then the stylized prefixes will be used to generate complete textual descriptions of 2 different styles accordingly. The two style encoders are actually two linear transformation matrices with different weights.

\begin{table}[t]
\tiny
\centering
\caption{Structure of the attribute and sentence branch. The bias of all linear layers is set to true. $n$ represents batch size. $c$ represents the channel size. $h$ and $w$ represent the height and width of the output feature map.}
\label{branch_structure}
\resizebox{0.5\linewidth}{!}{
\begin{tabular}{c|c|c}
\hline
Layers  &Paramters  & Output Size  \\
(CNN-MLP)  & (kernel,stride,pad)  & (n,c,h,w)  \\
\hline
Conv layer &(3,2,1)  & (n,c,h/2,w/2)   \\
Avgpool  &-& (n,c)\\
Linear& - & (n,768)\\
Dropout&rate=0.25&(n,768)\\
Linear& - & (n,2304)\\
Dropout&rate=0.25&(n,2304)\\
LeakyReLU&rate=1/5.5&(n,2304)\\
Rearange&-&(n,3,768)\\
\hline
Conv layers $\times$ 2&(3,2,1)  & (n,c,h/4,w/4)   \\
Avgpool  &-& (n,c)\\
Linear& - & (n,1536)\\
Dropout&rate=0.25&(n,1536)\\
Linear& - & (n,3840)\\
Dropout&rate=0.25&(n,3840)\\
LeakyReLU&rate=1/5.5&(n,3840)\\
Rearange&-&(n,5,768)\\
\hline
\end{tabular}}
\end{table}

\subsubsection{Details of Noise Filter Strategy}
To filter various noises in the LUPerson-NL~\cite{LUPnl} dataset, we propose Seed Filter Strategy, which is illustrated in Algorithm~\ref{Al1}. Specifically, it consists of three processes, each with its corresponding similarity threshold for filtering, reassigning or merging. 1) \textit{Filter-out}. For all samples with same identity, we cycle to calculate the similarity between the current sample and the center of other samples and exclude the samples whose similarity does not meet the filtering threshold $\sigma_{s}$ until the similarity of all samples with same identity meets it. 2) \textit{Reassignment}. According to the similarity and reassigning threshold $\sigma_{r}$, we reassign the excluded samples in the excluded dataset to the correct identity within a certain identity continual range. 3) \textit{Merger}. We merge the samples that should belong to the same identity but are divided into different identities according to the merging threshold $\sigma_{m}$. Through these processes, the samples with incorrect identity label annotations can be well filtered out or included in their expected identity set.

The following is a detailed explanation of the specific symbols in the Algorithm~\ref{Al1}. $S_{in}$ is the input dataset to be denoised while $S_{out}$ is the output denoised dataset. $S_{exclude}$ is the set of image samples to be excluded. $S_{merge}$ is the set of identities to be merged and it contains a lot of merged-identity pairs. $ID_k$ is the $k.th$ identity of the input dataset, with $n$ image samples $x_{i}^{k}$. $FID_k$ is the output final identity. $Sim$ is a similarity calculate function. If the variable it contained is a single identity, it will calculate the similarity between each sample and other samples' center. If it contains a image sample and a identity, it will calculate the similarity between the image sample and the center of all samples in the identity. If it contains two identities, it will calculate the similarity between the two centers of the two identities. $Merge$ is a function that merges the identities. We use cosine similarity to calculate the similarity between samples. For the hyper-parameters, $\sigma_s$ is set to 0.6. $\sigma_r$ is set to 0.65. $\sigma_m$ is set to 0.62. $r_a$ and $r_b$ are both set to 2.
        \vspace{-5mm}
        \begin{center}
        \begin{minipage}{0.6\linewidth}
        \begin{algorithm}[H]
        \caption{Seed Filter Strategy}\label{Al1}
        \begin{algorithmic}[1]
        \STATE Input $S_{in}=\{ID_k\}_{k=1}^N$, where $ID_k = \{x_i^k,\ldots,x_n^k\}$

        \FOR{$ID_k \in S_{in}$}

        \REPEAT
        
        \IF{$\min Sim(ID_k) < \sigma_s$}
        
        \STATE $x_j^k = \arg\min Sim(ID_k)$;
        
        \STATE $ID_k = ID_k-x_j^k,x_j^k \Rightarrow S_{exclude};$
        
        \ENDIF
        
        \UNTIL $\min Sim(ID_k) \geq \sigma_s$
        
        \ENDFOR
        
        \FOR{$x_j^k \in S_{exclude}$}
        
        \IF{$\max \{Sim(x_j^k,ID_i)\}_{i=k-r_a}^{k+r_a} \geq \sigma_r$}
        
        \STATE $ID_r = \arg\max\{Sim(x_j^k,ID_i)\}_{i=k-r_a}^{k+r_a}$;
        
        \STATE $ID_r = ID_r + x_j^k $;
        
        \ENDIF
        
        \ENDFOR
        
        \FOR{$ID_k \in \{ID_i\}_{i=1}^{N}$}
        
        \IF{$\{Sim(ID_k,ID_{k+i})\}_{i=1}^{r_b}>\sigma_m$}
        
        \STATE $\{ID_k,ID_{k+i}\} \Rightarrow S_{merge}$
        
        \ENDIF
        
        \ENDFOR
        
        \FOR{$FID_i \in S_{merge}$}
        
        \FOR{$ FID_j \in \{FID_j\}_{j=i+1}^{i+r_b}$}
        
        \IF{$FID_i \cap FID_j \neq \varnothing$}
        
        \STATE $FID_i = Merge(FID_i,FID_j)$
        
        \ENDIF
        
        \ENDFOR
        
        \STATE $FID_i \Rightarrow S_{out}$
        
        \ENDFOR
        
        \STATE Output $S_{out}$;
        
        \end{algorithmic}
        \end{algorithm}
        \end{minipage}
        \end{center}

\vspace{-2mm}
\subsubsection{Details of Data Distribution Strategy}
\vspace{-2mm}
For the image part, there are some bad pictures with blocking, blurring and multiple people. Due to the poor quality, these pictures are not a good choice for training. Also, the number of images per identity in LUPerson-NL~\cite{LUPnl} is very unbalanced. Some identities have thousands of images, while others have only few images. To further improve the quality of our dataset, we have adopted the following strategies: 1) \textit{Ensure the qualities of generated attributes.} In our setting, the attributes generated by poor quality pictures will contain a large number of unknowns. To filter these poor quality pictures, the pictures with three or more unknown attributes will be directly filtered. Particularly, if the upbody and downbody attributes of a picture are unknown simultaneously, it will also be filtered. 2) \textit{Ensure the consistency of gender attribute.} Ensuring the gender consistency of an id plays a role in promoting the stability of the model training. However, for those pictures with poor quality, it is often very challenging to correctly predict their gender attributes. So we adopt a gender voting mechanism. That is, for all pictures in an identity, if the gender attribute with the maximum frequency of occurrence is greater than 0.7, the gender of all pictures in the identity will be automatically changed to this gender. Otherwise, the pictures in the identity will be considered as bad pictures, and will be filtered out. 3) \textit{Balance the distribution of identities.} If the number of images of an identity is less than the minimum number of 5, this identity will be filtered. If the number of images of an identity is greater than the maximum number of 20, randomly select 20 images from all images of the identity and filter out the others.

\begin{wrapfigure}[19]{r}{7cm}
\vspace{-5mm}
\centering
	\includegraphics[width=\linewidth]{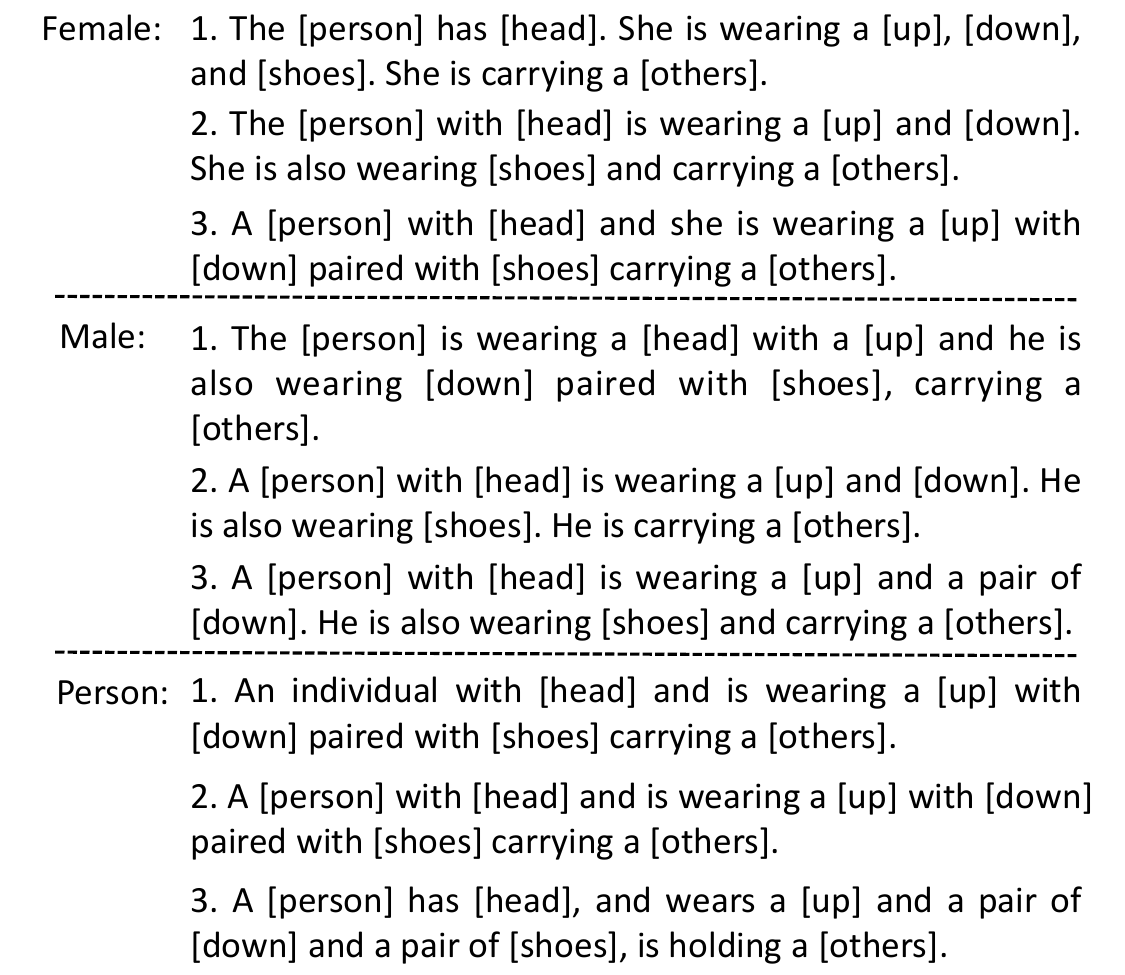}
\caption{Some examples in the mask standard sentence library for the attributes-all-known situation.}
\label{fig:library}
\end{wrapfigure}
For the text part, we aim to generate three sentences for an image. Two of them are directly generated by SPAC while the another is generated by Attributes Prompt Engineering exploiting the attribute descriptions generated by SPAC. In many processes of our approach, we must extract the attribute phrases from a textual description. For example, in order to guild the SPAC to generate attribute descriptions, we have to extract the attribute labels from the paired textual descriptions. The extracting process is as the following: 1) \textit{Construct the keywords set for each attribute.} Firstly, we extract all the nouns in the datasets'~\cite{textreid,SSAN} captions by nltk tools and sort them by frequency. Then, for each attribute, we manually select from the nouns and construct the keywords set. 2) \textit{Assign the noun phrases to relevant attributes.} For a textual description, we firstly extract the noun phrases by nltk tool. Then, according to the keywords set, we assign each noun phrase to the correct attribute. On this way, we finally finish extracting the attribute phrases from a a textual description. For Attributes Prompt Engineering, as shown in Fig.~\ref{fig:library}, we have created a mask standard sentence library with many diversities. To construct the library, by exploiting the nltk tool and the above attributes extracting method, we first mask all attributes phrases of the captions in CUHK-PEDES and ICFG-PEDES. Then We manually select sentences with good sentence structure from them to form the mask standard sentence library. The library totally consists of 1664 sentences for different generated attribute missing situations.

\subsection{Experimental Setup}
\subsubsection{Implementation Details}
\label{implementationdetails}
 \textbf{During the training of SPAC,} we adopt ResNet-101~\cite{resnet} together with a Feature Pyramid Network (FPN)~\cite{fpn} as the prefix encoder and GPT2~\cite{gpt2} as the generator. The ResNet-101 and GPT2 are both pre-trained on their respective pre-training tasks. All images are resized to 384 × 128 and normalized with mean and std of [0.485, 0.456, 0.406], [0.229, 0.224, 0.225], which are calculated from all images in ImageNet. We use the combination of CUHK-PEDES~\cite{textreid} and ICFG-PEDES~\cite{SSAN} as the training dataset. We adopt horizontally flipping to augment data, where each image has 50\% probability to flip randomly. We aim to generate the textual descriptions with two style type. The learning rate is fixed at 0.0001. The balance factor $\lambda$ is set as 0.15. There are six types of attributes including gender, head, upper body, lower body, shoes and belongings. The prefix lengths of attributes and relation are 3 and 5 respectively. We train it on 4 × Geforce 3090 GPUs for 30 epochs with a batch size of 64 totally, which takes approximately 1.6 days. The optimizer is AdamW~\cite{adamw} with the default setting.

\noindent
\textbf{During the training of PLIP,} we adopt four types of backbone as the visual encoder, \ie ResNet50, ResNet101, ResNet152 and Swin Transformer Base. The pre-trained BERT~\cite{bert} is utilized as the textual encoder and we only unfreeze the last 5 layers, keeping other parameters frozen. All images are resized to 256 × 128 and normalized with mean and std of [0.357, 0.323, 0.328], [0.252, 0.242, 0.239], which are calculated from our proposed SYNTH-PEDES. We adopt horizontally flipping to augment data, where each image has 50\% probability to flip randomly. For ResNet50, we train our model on 4 × Geforce 3090 GPUs for 70 epochs with a batch size of 512 totally, which takes approximately 15.2 days. The base learning rate is set to 0.002 and decreased by 0.1 at the epoch of 30 and 50. Besides, the learning rate warm-up strategy is adopted in the first 10 epochs. The learning rate of BERT has a 0.1 decay. For other types of visual encoder, there are some differences on the learning rate and batch size setting. The hyper-parameters in the objective function are set to $\lambda_1=0.02$ and $\lambda_2=0.1$. The optimizer is Adan~\cite{adan} with the default setting. We adopt the mixed precision training mode by Apex.

\noindent
 \textbf{For each person-centric downstream task,} we reproduce a range of state-of-the art methods as the baselines. If no special instructions are given, we perform the experiments by just replacing the backbone in each baseline to the models pre-trained by our proposed method. Meanwhile, for text-based person Re-ID, we adopt the standard metrics Rank-\textit{k} (\textit{k}=1,5,10) to evaluate the model performance. For image-based person Re-ID and person search, we follow the popular evaluation metrics: the mean Average Precision (mAP) and the Rank-\textit{k} (\textit{k}=1,5,10). For person attribute recognition, we adopt four evaluation metrics including accuracy (Acc), mean accuracy (mA), recall (Rec) and F1 value (F1). For human parsing, we use mean accuracy (mA) and mean intersection over union (mIoU) as the evaluation metrics.

\vspace{-2mm}
\subsubsection{Benchmarks} 
\vspace{-1mm}
To evaluate our proposed approach, we perform in-depth experiments on eleven datasets for five downstream person-centric tasks. For text-based person Re-ID, we conduct extensive experiments on two popular pubic datasets, \ie CUHK-PEDES~\cite{textreid} and ICFG-PEDES~\cite{SSAN}. For image-based person Re-ID, we use three popular public datasets, \ie Market1501~\cite{market}, MSMT17~\cite{MSMT17} and DukeMTMC~\cite{dukemtmc}. For person attribute recognition, three large-scale datasets PETA~\cite{peta}, PA-100K~\cite{a2} and RAP~\cite{rap} are used. For person search, we adopt PRW~\cite{PRW} and CUHK-SYSU~\cite{CUHK-SYSU} in our experiments. For human parsing, the LIP dataset~\cite{LIP} and PASCAL-Person-Part dataset~\cite{PASCAL} are used. Next is an introduction to representative datasets in each task.

CUHK-PEDES dataset~\cite{textreid} is the first and most commonly used benchmark for text-based person re-identification. It contains 40,206 images and 80,412 textual descriptions for 13,003 identities. Each person image has two corresponding textual descriptions on average. We dopt the same data split as~\cite{textreid}. The training set has 34,054 images of 11,003 identities. The validation and test set have 3,078 and 3,074 images of 1,000 identities, respectively.

Market1501 dataset~\cite{market} includes 32,668 images of 1,501 persons captured from 6 different cameras. The DPM detector is employed to crop the bounding boxes for these pedestrians. The dataset is divided into a training set comprising 12,936 images of 751 persons and a test set with 3,368 query images and 19,732 gallery images of 750 persons.

PA-100K dataset~\cite{a2} contains 100,000 pedestrian images with resolutions ranging from $50 \times 100$ to $758 \times 454$ captured from 598 different cameras. Each pedestrian image is annotated with 26 commonly used attributes. As the official protocol, there are 80,000 images for training, 10,000 images for validation, and 10,000 images for testing.

PRW dataset~\cite{PRW} is a widely used dataset for person search which contains 11,816 video frames captured by 6 different cameras, and 34,304 manually annotated bounding boxes. All people are divided into labeled identities and other unknown ones. The training set comprises 5,704 images and 482 identities. The test set contains 6,112 images and 2057 query people.

LIP dataset~\cite{LIP} is a large human parsing dataset containing 50,462 images with manual pixel-wise annotations for 19 semantic human parts. The dataset is collected from the real-world scenarios. It contains human images with various appearances, heavily occlusions and low-resolutions, which introduces many challenges. LIP is divided into 30,462 images for train set, 10,000 images for validation set and 10,000 images for test set. 

\vspace{-2mm}
\subsection{Thorough Ablation Studies and Analyses}
\label{throughablation}
\vspace{-1mm}
In this section, we perform in-depth ablation studies to analyze the effectiveness of each part of our work. If not specially stated, we use ResNet-50 as the visual encoder and pre-train on the sub-dataset of SYNTH-PEDES, which has 10,000 identities, 139,564 images and 353,617 textual descriptions. The evaluation is mainly performed on the benchmarks of text-based person Re-ID and image-based person Re-ID using the zero-shot evaluation protocol. 

\begin{wraptable}[6]{r}{7cm}
\vspace{-12mm}
\large
\centering
\caption{Ablation study on the impact of each pretext task, all using default settings.}
\label{tab9.1}
\resizebox{\linewidth}{!}{
\begin{tabular}{c|ccc|ccc|ccc}
\hline 
\multicolumn{1}{c|}{\multirow{2}{*}{\#}} & \multicolumn{3}{c|}{Components}&\multicolumn{3}{c|}{CUHK-PEDES}&\multicolumn{3}{c}{Market1501}\\
\cline{2-10}
& IVLC &TIC &IAP&R@1&R@5&R@10&R@1&R@5&R@10    \\
\hline 

1 &\checkmark& &&30.2	&53.3&64.0	&62.3&79.9	&85.7 \\
2 &\checkmark& \checkmark& &31.4	&55.2&65.9	&62.9&80.7	&86.2 \\
3&\checkmark& &\checkmark&30.5	&54.4&64.3	&62.7&80.5	&86.1 \\
4&&\checkmark&\checkmark&-&-&-&39.3	&61.4&70.4	 \\
\rowcolor{gray!20}
5&\checkmark&\checkmark&\checkmark&\textbf{32.5}	&\textbf{56.3}&\textbf{66.6}	&\textbf{63.1}&\textbf{80.8}	&\textbf{86.3} \\
\hline 
\end{tabular}}
\end{wraptable}

\subsubsection{Effectiveness of Each Component}
\textbf{Impact of each pretext task.} We have designed three pretext tasks to implement pre-training. To assess the impact of pretext tasks on the generalizability of learned representations, we directly evaluate the performance of pre-trained models on the downstream datasets without fine-tuning. As indicated in Tab.~\ref{tab9.1}, each task contributes to the model's zero-shot capability on CUHK-PEDES and Market1501. Combining all tasks leads to the optimal performance, highlighting the significant role each task plays in facilitating the learning of generic person representations.

\begin{wrapfigure}[13]{r}{7cm}
\vspace{-5mm}
\centering
	\includegraphics[width=\linewidth]{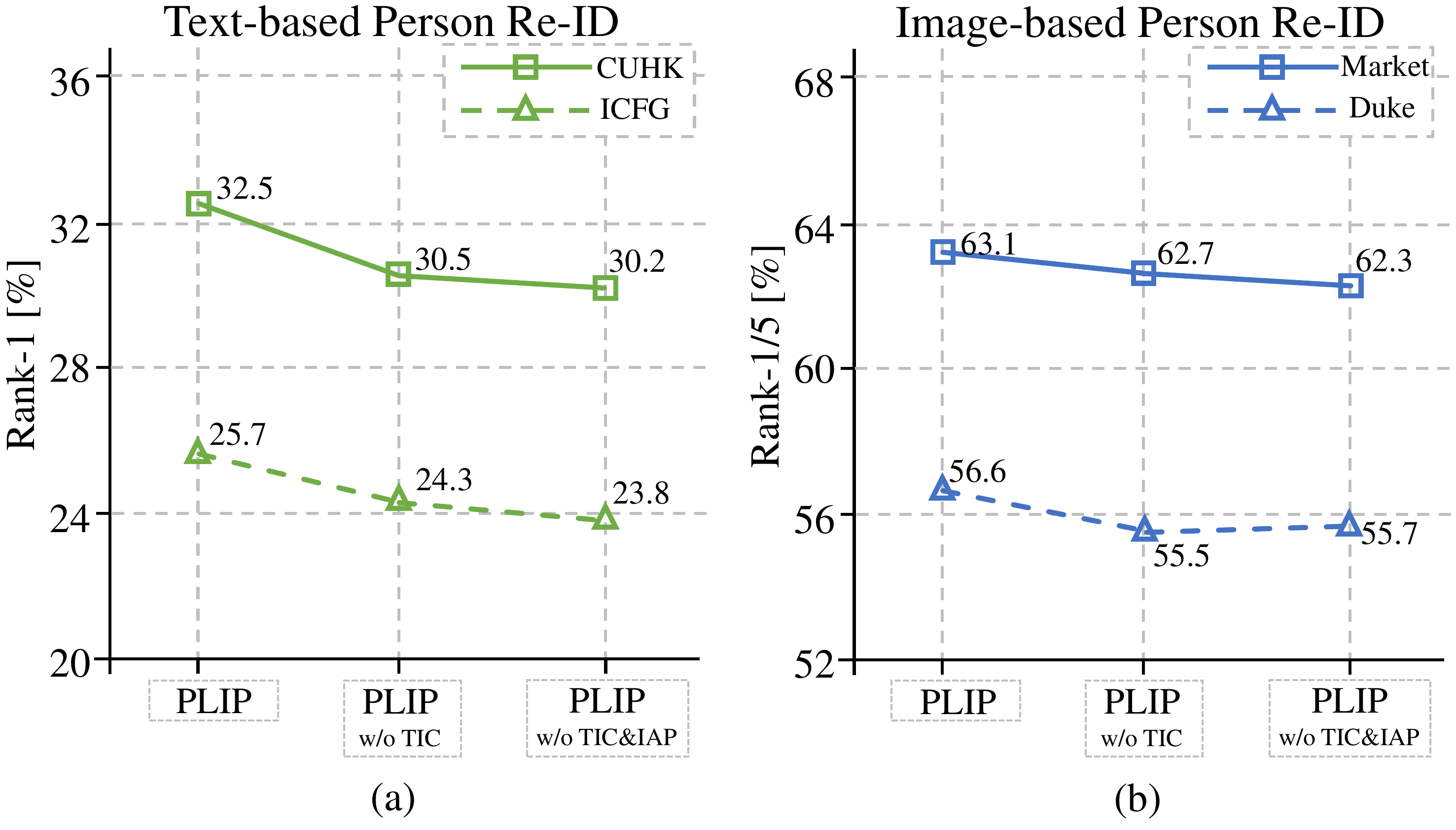}
 \vspace{-5mm}
\caption{The effectiveness of TIC on transfer capability. TIC brings significant improvements.}
\label{sic_matters}
\end{wrapfigure}
\textbf{TIC is essential for transfer capability.} After pre-training with different settings, we evaluate the model's performance on four datasets under the zero-shot setting. The metric on DukeMTMC~\cite{dukemtmc} is rank-5, while others~\cite{textreid,SSAN,market} are all rank-1. As shown in Fig.~\ref{sic_matters}, without TIC, the learned representation exhibits noticeably weaker performance on downstream datasets, averaging a 2.45\% decrease compared to the default method on CUHK-PEDES and Market1501. However, the introduction of TIC significantly improves the transfer performance. We believe this is because TIC encourages the model to comprehend textual semantics and accomplish body part localization and coloring, which holds significant implications for learning generic representations.

\begin{wrapfigure}[19]{r}{7cm}
\vspace{-5mm}
\centering 
	\includegraphics[width=0.8\linewidth]{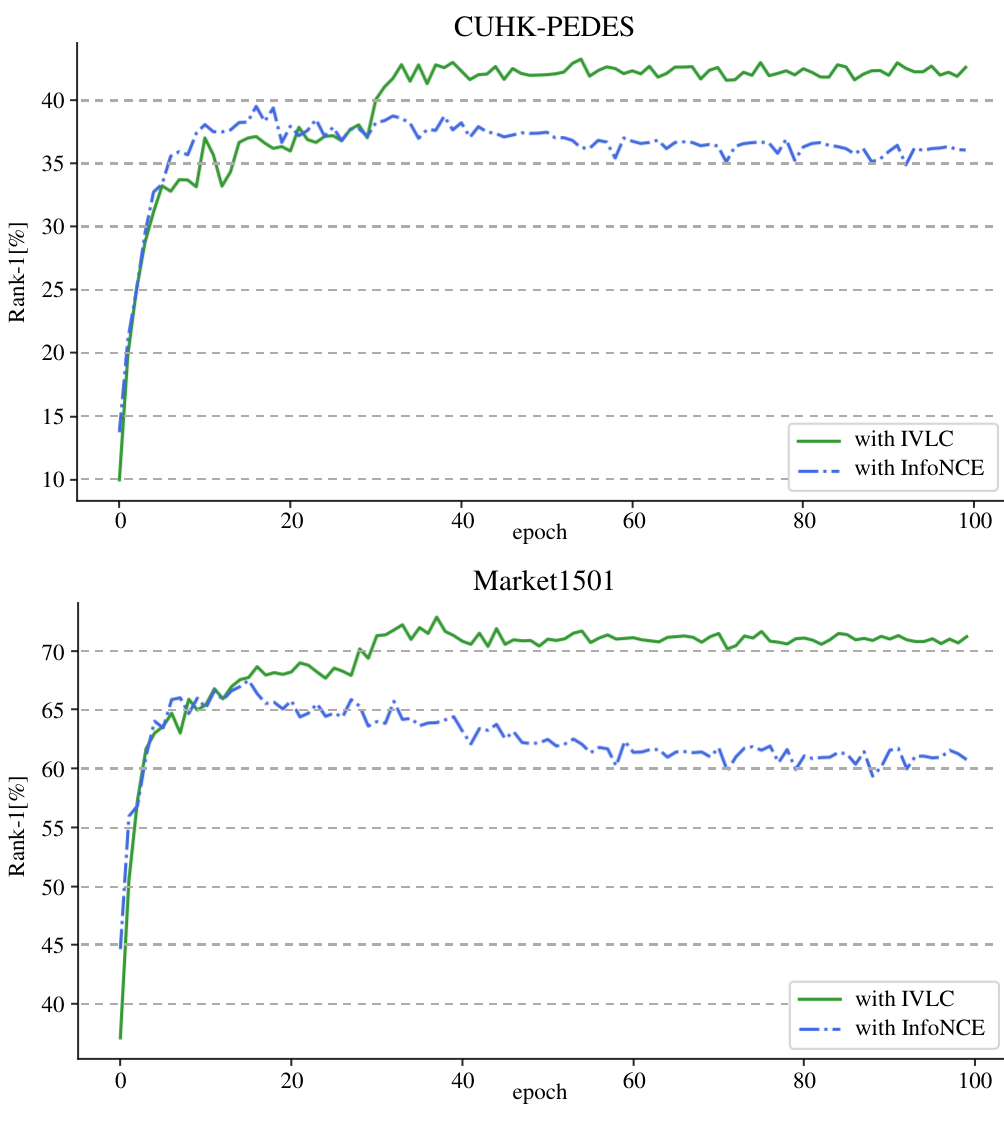}
\caption{The effectiveness of IVLC in stable representation learning, which brings much better performance.}
\label{fig:ivlcstable}
\end{wrapfigure}
\textbf{IVLC contributes to stable representation learning.}
 We utilize the IVLC loss to optimize the model at identity-level. We conduct an experiment to verify the effectiveness of IVLC in stable and meaningful representation learning. The training samples in this experiment are the first one million images from SYNTH-PEDES. We test the zero-shot performance of the trained models with different settings on the CUHK-PEDES and Market1501 datasets. As the results shown in Fig.~\ref{fig:ivlcstable}, the model trained with instance-level InfoNCE loss significantly underperforms compared to the model trained with identity-level IVLC loss. Additionally, as the training process extends, the model trained with InfoNCE loss experiences a notable decline in performance, whereas the performance of the model trained with IVLC loss tends to stabilize. These results demonstrate that IVLC contributes to stable and better representation learning.

\vspace{-3mm}
\subsubsection{Dataset Quality Evaluation}\label{datasetqualityeval}
The quality of dataset is critical for large-scale representation learning. We have conducted quantitative experiments to evaluate the quality of our dataset compared to three manual annotated datasets~\cite{textreid,SSAN,RSTPReid} and an existing synthetic dataset~\cite{VTBR}.

\textbf{Manual Evaluation}

\begin{wraptable}[10]{r}{7cm}
\vspace{-4.5mm}
\centering
\caption{Manual evaluation results. Our SYNTH-PEDES dataset is competitive with manual datasets and surpasses an existing synthetic dataset by a large margin. }
\vspace{-2mm}
\label{manualeval}
\resizebox{\linewidth}{!}{
\begin{tabular}{c|ccccc|c}
\hline
 \multicolumn{1}{c|}{\multirow{2}{*}{Dataset}}  & \multicolumn{5}{c}{Score}  &  \multicolumn{1}{|c}{\multirow{2}{*}{Average}} \\
 \cline{2-6}
  & 1 & 2 & 3 & 4&5& \\
\hline   
CUHK-PEDES~\cite{textreid}&14&24&146& 196&120&3.77\\  
ICFG-PEDES~\cite{SSAN}&12&44&124&150&170&3.84\\ 
RSTPReid~\cite{RSTPReid}&11&26&149&146&168&3.87\\ 
\hline  
FineGPR-C~\cite{VTBR}&24&41&435&0&0&2.82\\ 
\rowcolor{gray!20}
\textbf{SYNTH-PEDES}&18&30&162&132&158&3.76\\ 
\hline
\end{tabular}
}
\end{wraptable}
 We randomly select 500 image-text pairs from our SYNTH-PEDES dataset and other manually annotated or synthetic datasets, forming 2500 image-text pairs to be evaluated. The score of the evaluation is divided into five levels. The first is that the most part of the description of the image is incorrect. The second is that a small part of the description of the image is incorrect. The third is that the description is correct but rough. The fourth is that the description is correct and generally detailed. The fifth is that the description is completely correct and very detailed. As shown in Tab.~\ref{manualeval}, our SYNTH-PEDES dataset is competitive in quality compared to these manually annotated dataset at about 200 times the amount of image-text pairs.

\textbf{Automatic Evaluation}

\begin{wrapfigure}[19]{r}{7cm}
\vspace{-5mm}
\centering 
	\includegraphics[width=\linewidth]{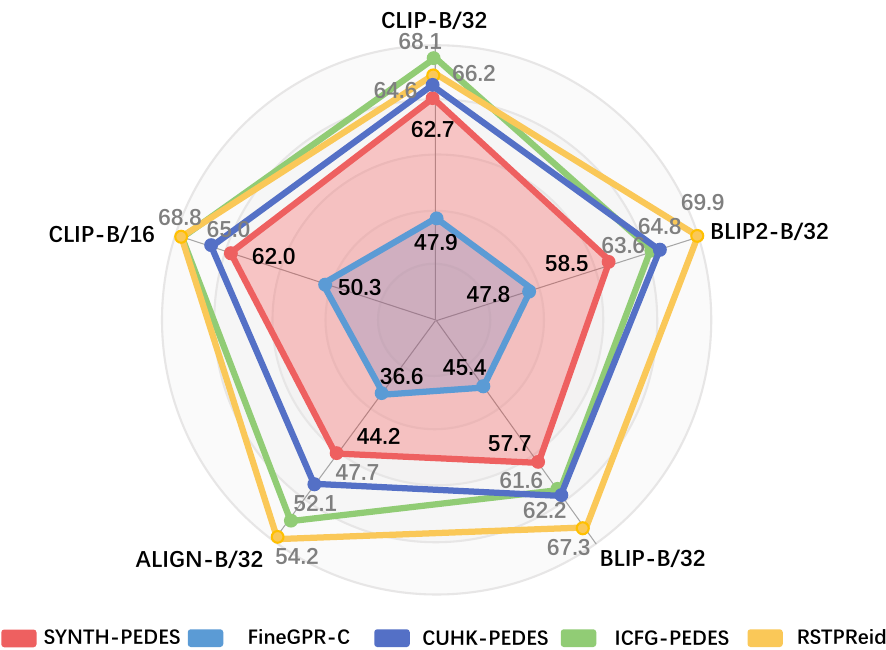}
\caption{Automatic evaluation results by 5 pre-trained models. Our dataset are about 200 times larger than manually annotated datasets, yet still mantains competitive data quality.}
\label{fig:autoeval}
\end{wrapfigure}
We utilize some multi-modal pre-trained models like CLIP~\cite{CLIP}, ALIGN~\cite{Align}, BLIP~\cite{BLIP} and BLIP2~\cite{BLIP2} to calculate the cosine similarity of image-text pairs in each dataset. Moreover, considering the varied sensitivity of each model to image-text similarity, we employ the following normalization to ensure that the results are approximately within the same range. For each model's evaluation result, we initially identify its maximum similarity across all datasets and assign it a score of 100 points. Subsequently, the score for each dataset is determined as the average similarity relative to this highest similarity. As shown in Fig.~\ref{fig:autoeval}, under the evaluation of five models, our SYNTH-PEDES dataset not only gains competitive scores compared to manually annotated datasets, but also surpasses the existing synthetic dataset~\cite{VTBR} by a large margin.

\subsubsection{Different Pre-training Settings}
\begin{table*}[h]
\centering
\begin{minipage}{\linewidth}
\subfloat[Different caption methods.
]{
\centering
\tablestyle{1pt}{1}
\begin{tabular}{c|cccccc}
 \multicolumn{1}{c|}{\multirow{2}{*}{Downstream}}  & \multicolumn{2}{c}{SAT} &  \multicolumn{2}{c}{ClipCap} &  \multicolumn{2}{c}{\cellcolor{gray!20}SPAC} \\
  & R@1 &R@10& R@1 & R@10 &R@1& R@10\\
\hline   
CUHK-PEDE& 23.6  &  55.8& 24.3  &  56.2& \textbf{26.6}  &  \textbf{59.8} \\     
ICFG-PEDES& 17.9  &  48.8& 18.4  &  49.2& \textbf{20.9}  &  \textbf{52.3} \\    
RSTPReid& 16.2  &  51.0& 16.5  &  51.4& \textbf{18.5}  & \textbf{ 54.4} \\    
\hline
Market1501& 57.3  &  81.8& 57.4  &  81.8& \textbf{58.4}  &  \textbf{82.8} \\   
DukeMTMC& 37.8  &  59.4& 38.2  &  59.6& \textbf{39.3}  &  \textbf{61.0} \\    
MSMT17& 16.9  & 36.2& 17.1  &  36.3& \textbf{18.3}  &  \textbf{37.9} \\   
\end{tabular}
}
\subfloat[Different dataset quality.
]{
\centering
\footnotesize
\tablestyle{1pt}{1}
\begin{tabular}{c|cccccc}
 \multicolumn{1}{c|}{\multirow{2}{*}{Downstream}}  & \multicolumn{3}{c}{Low-quality}  &  \multicolumn{3}{c}{\cellcolor{gray!20}High-quality} \\
  & R@1 &R@5& R@10 & R@1 &R@5& R@10\\
\hline   
CUHK-PEDE& 28.3  &  51.7& 62.4  &  \textbf{29.2}& \textbf{51.9}  &  \textbf{62.9} \\  
ICFG-PEDES& 22.2  &  43.6& 53.4  &  \textbf{23.8}& \textbf{44.9}  & \textbf{55.5}  \\  
RSTPReid& 19.3  &  41.7& 54.7  &  \textbf{20.0}& \textbf{42.5}  &  \textbf{55.2} \\  
\hline
Market1501& 57.9  &  77.2& 83.3  & \textbf{58.6} & \textbf{77.4}  & \textbf{83.4}  \\   
DukeMTMC& 39.6  &  56.0& 62.8  &\textbf{40.9}  & \textbf{57.2}  &\textbf{64.4}   \\    
MSMT17& 19.3  &  32.6& 39.1  &  \textbf{19.6}& \textbf{33.0}  &  \textbf{39.3} \\    
\end{tabular}
}
\end{minipage}
\caption{Ablation studies on caption methods in (a) and dataset quality in (b).}
\vspace{-2mm}
\label{table:dataset}
\end{table*}

\begin{table*}[h]
\centering
\begin{minipage}{\linewidth}
\subfloat[Different training objectives.
]{
\hspace{-7mm}
\centering
\tablestyle{0.5pt}{1}
\begin{tabular}{cc|cccc}
\multicolumn{1}{c}{\multirow{2}{*}{IVLC}} & \multicolumn{1}{c|}{\multirow{2}{*}{TIC}}  & \multicolumn{2}{c}{CUHK}  &  \multicolumn{2}{c}{Market} \\
 & & R@1 & R@10 & R@1 & R@10\\
\hline   
$\mathcal{L}_{sd.}$&Manh.& 18.7  &  49.0& 46.4  &  73.8 \\
$\mathcal{L}_{al.}$&Manh.& 17.6  &  45.4& 44.0  &  71.9 \\
$\mathcal{L}_{cm.}$&Manh.& 19.1  &  48.5& 44.4  &  75.4 \\ 
\hline  
$\mathcal{L}_{sd.}$&Eucl.& 30.0  &  63.5& 63.1  &  86.1 \\  
$\mathcal{L}_{al.}$&Eucl.& 31.8  &  64.1& 63.0  &  85.7 \\ \rowcolor{gray!20} 
$\mathcal{L}_{cm.}$&Eucl.& \textbf{32.5}  &   \textbf{66.6}&  \textbf{63.1}  &   \textbf{86.3} \\
\end{tabular}
}
\subfloat[Different prediction difficulty.
]{
\centering
\footnotesize
\tablestyle{2pt}{1}
\begin{tabular}{c|cccc}
 \multicolumn{1}{c|}{\multirow{2}{*}{$G(x)$}}  & \multicolumn{2}{c}{CUHK}  &  \multicolumn{2}{c}{Market} \\
  & R@1 & R@10 & R@1 & R@10\\
\hline   
0.2&32.1&  65.9& 63.2 &  86.6 \\  
0.5& 30.9  & 65.3& 62.9 & 85.2 \\ 
0.8& 31.6  &  66.1& 62.7  &  86.9 \\ 
\hline  
$x^2$& 32.3  &  66.0& 63.4  &  85.9 \\  
 \rowcolor{gray!20}
$\sqrt{x}$& \textbf{32.5}  &  \textbf{66.6}& 63.1  &  86.3 \\  
$x$& 31.7 & 65.5& \textbf{63.7}  & \textbf{86.5} \\
\end{tabular}
}
\subfloat[Different pooling methods.
]{
\centering
\footnotesize
\tablestyle{0.5pt}{1}
\begin{tabular}{cc|cccc}
\multicolumn{1}{c}{\multirow{2}{*}{Visual}} & \multicolumn{1}{c|}{\multirow{2}{*}{Textual}}  & \multicolumn{2}{c}{CUHK}  &  \multicolumn{2}{c}{Market} \\
 & & R@1 & R@10 & R@1 & R@10\\
\hline   
Aver.&Pooler& 26.6  &  60.8& 55.5  &  81.0 \\  
Max.&Pooler.& 26.2  &  59.5& 56.2  &  83.0 \\
Attn.&Pooler.& 24.3  &  57.1& 57.3  &  81.9 \\ 
\hline  
Aver.&Aver.& 30.2  &  65.6& 60.1 &  84.1 \\  \rowcolor{gray!20}
Max.&Aver.&  \textbf{32.5}  &   \textbf{66.6}&  \textbf{63.1}  &   \textbf{86.3} \\
Attn.&Aver.& 27.9  &  62.4& 61.3  &  84.9 \\ 
\end{tabular}
}
\end{minipage}
\hspace{-4mm}

\caption{Ablation studies on training objectives in (a), prediction difficulty functions in (b) and pooling methods in (c).}
\vspace{-5mm}
\label{table:ablation_overall}
\end{table*}

\textbf{Pre-training with other caption methods.} We have compared our proposed SPAC with the previous representative caption methods SAT~\cite{SAT} and ClipCap~\cite{clipcap} in generating captions for person images. The compared methods generate only one caption per image, lacking diversity. To ensure a fair comparison, we exclusively utilize SPAC to generate a single caption for the sub-dataset derived from SYNTH-PEDES. Totally, for each method, there are 139,564 image-text pairs for pre-training. As shown in Tab.~\ref{table:dataset} (a), the performance of SPAC on six datasets is much better than the others, showing the superiority of our SPAC.

\textbf{High-quality dataset leads to good pre-training.}
In this experiment, we consider the original LUPerson-NL dataset~\cite{LUPnl} and our dataset as low-quality and high-quality dataset, respectively. We conduct a performance comparison on downstream tasks following pre-training on each dataset. To ensure parity, an equal number of identities and image samples are randomly selected for each dataset, and captions are generated using SPAC. The pre-training process involved a total of 214,053 image-text pairs for each dataset. As shown in Tab.~\ref{table:dataset} (b), the downstream performance of the low-quality dataset is markedly inferior to ours. These results collectively validate the efficacy of the strategies employed in our dataset gathering process.

\begin{wrapfigure}[12]{r}{7cm}
\vspace{-4mm}
\centering
	\includegraphics[width=\linewidth]{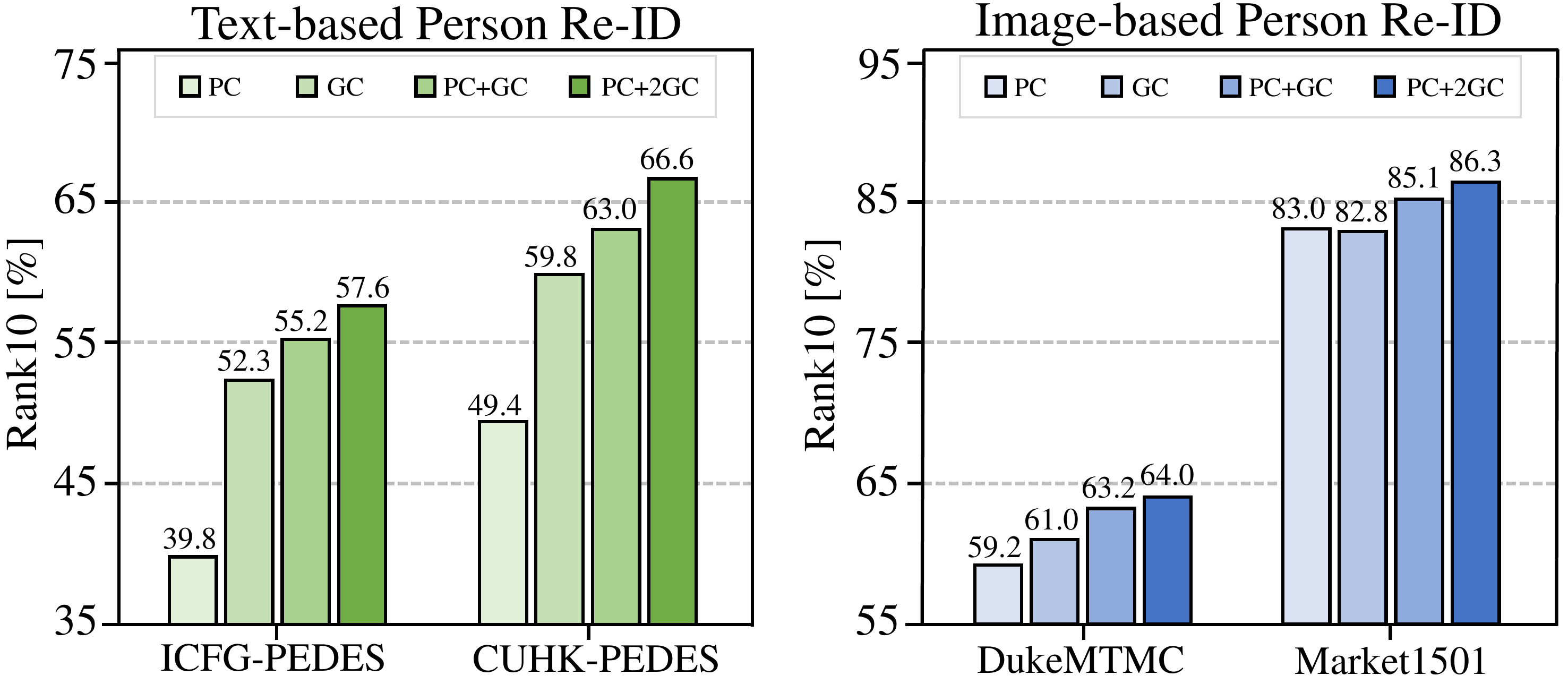}
 \vspace{-5mm}
\caption{The diversity of textual descriptions matters. PC and GC mean prompt caption and generate caption, respectively.}
\vspace{-5mm}
\label{diversity_matters2}
\end{wrapfigure}
\textbf{The diversity of textual descriptions matters.}
Our dataset has textual descriptions with different styles for each image. To validate the effectiveness of this diversity, we assess it through pre-training with datasets exhibiting varying degrees of textual diversity, followed by a comprehensive generalizability study on downstream tasks. Specifically, we have studied four different degrees of textual diversity from weak to strong, while keeping the number of person images and identities consistent. As shown in Fig.~\ref{diversity_matters2}, the fourth case, with the highest degree of textual diversity, has the best performance. Also, compared to the first case, the superior performance observed in the second case further substantiates the significance of diversity, as the prompt caption exhibits less diversity than the generated caption.

\subsubsection{Optional Practices}
\textbf{Training Objectives of IVLC and TIC.}
The different combinations and implementations of the loss functions for IVLC and TIC are shown in Tab.~\ref{table:ablation_overall} (a). For the IVLC task, we have implemented three identity-based loss functions for vision-language contrastive learning. Specifically, $\mathcal{L}_{al.}$ represents the alignment loss proposed in~\cite{vitaa}, $\mathcal{L}_{sd.}$ represents the SDM loss proposed in~\cite{IRRA}, and $\mathcal{L}_{cm.}$ represents the CMPM loss proposed in~\cite{CMPM}. For the TIC task, to measure the error between the color-restored image and the original colorful image, we tried using Euclidean distance and Manhattan distance. In fact, any function meeting the requirements can be used in our learning framework. However, as shown in Tab.~\ref{table:ablation_overall} (a), the combination of CMPM loss and Euclidean distance leads to the best performance, and gains rank-1 32.5\% and 63.1\% on each dataset accordingly. 

\textbf{Different prediction difficulty.}
Curriculum learning~\cite{curriculum,HICO} demonstrates that it could be useful to organize the training samples from easy to hard. It is inspired by the human learning process and has achieved success in a wide range of tasks. To explore the effectiveness of this, we consider the masked textual descriptions with different masking probabilities as samples with different difficulty levels. Specifically, we define $G(x)$ as the difficulty function and explore a variety of optional functions, including constant, $x$, $x^2$, and $\sqrt{x}$. By using the gradually increasing functions, we start the $G(x)$ grows from 0.2 to 0.8, which means the training samples will change from easy to hard at different rates. As shown in Tab.~\ref{table:ablation_overall} (b), the results suggest that satisfactory performance on both datasets can be achieved with $\sqrt{x}$ in pre-training. This means the gradual difficulty sample strategy truly plays a role in encouraging the model to learn generic person representations better.

\textbf{Pooling methods of global embeddings.}
We have studied the impact of different pooling methods of global embeddings. In the textual branch, we investigated two methods for obtaining the global embedding, namely pooler-out and average pooling. Additionally, in the visual branch, we explored three pooling operations on the last stage feature: average pooling, maximum pooling, and attention-based pooling. As shown in Tab.~\ref{table:ablation_overall} (c), the combination of maximum pooling and average pooling consistently yields the best performance on both datasets, establishing it as the default setting.

\begin{wraptable}[11]{r}{7cm}
\vspace{-4mm}
\centering
\caption{Comparable results of different hyper-parameters in overall objective function. We show the best score in bold.}
\label{lossweights}
\resizebox{\linewidth}{!}{
\begin{tabular}{c|ccc|ccc}
\hline
\multicolumn{1}{c|}{\multirow{2}{*}{Settings}} & \multicolumn{3}{c|}{CUHK-PEDES}&\multicolumn{3}{c}{Market1501}\\
\cline{2-7}
 & R@1 & R@5 & R@10 & R@1 & R@5 & R@10  \\
\hline
$\lambda_{1}=1,\lambda_{2}=1$ &28.7 &51.9 &62.8 &57.6 &77.1 &83.4\\
$\lambda_{1}=0.2,\lambda_{2}=1$ &30.3 &53.6 &64.1 &61.7 &79.2 &85.5\\

$\lambda_{1}=0.05,\lambda_{2}=1$ &31.8 &55.1 &65.4 &62.2 &79.9 &85.6\\
$\lambda_{1}=0.02,\lambda_{2}=1$ &32.3 &55.8 &65.9&62.6 &80.3 &85.9\\
$\lambda_{1}=0.01,\lambda_{2}=1$ &31.7 &55.3 &65.1 &62.4 &80.1 &85.7\\
$\lambda_{1}=0.02,\lambda_{2}=0.5$ &32.5 &55.9 &66.1 &62.9 &80.9 &85.9\\
$\lambda_{1}=0.02,\lambda_{2}=0.2$ &32.3 &56.1 &66.4 &62.7 &\textbf{81.1} &86.1\\
 \rowcolor{gray!20}
$\lambda_{1}=0.02,\lambda_{2}=0.1$ &\textbf{32.5} &\textbf{56.3} &\textbf{66.6} &\textbf{63.1} &80.8 &\textbf{86.3}\\
\hline
\end{tabular}}
\end{wraptable}
\textbf{Effectiveness of different hyper-parameters.}
Since our objective function consists of three losses and each loss is critical to the overall performance, we further search for the optimal hyper-parameters to balance the loss weights. As shown in Tab.~\ref{lossweights}, PLIP achieves the best performance when $\lambda_1=0.02$ and $\lambda_2=0.1$, surpassing the version of $\lambda_{1,2}=1$ by 3.8\% and 5.5\% rank-1 on CUHK-PEDES and Market1501, respectively. This is mainly attributed to the loss values of three pretext tasks. The IVLC loss is the primary component of the objective function, aiming to achieve the visual and textual modality association. The TIC and IAP losses are employed to better assist in representation learning, and their loss values are relatively large. Therefore, adjusting the weights reasonably is also crucial for performance.

\subsubsection{Learning with Other Pre-training Methods}
\begin{wraptable}[11]{r}{7cm}
\vspace{-4mm}
\centering
\caption{Comparable results of PLIP with other representative pre-training methods. We show the best score in bold.}
\label{otherpretrain}
\resizebox{\linewidth}{!}{
\begin{tabular}{c|c|ccc|ccc}
\hline
\multicolumn{1}{c|}{\multirow{2}{*}{Method}} &\multicolumn{1}{c|}{\multirow{2}{*}{Backbone}}& \multicolumn{3}{c|}{CUHK-PEDES}&\multicolumn{3}{c}{Market1501}\\
\cline{3-8}
 & &R@1 & R@5 & R@10 & R@1 & R@5 & R@10  \\
\hline
CLIP &RN50&38.1 &62.6 &73.0 &70.1 &84.7 &89.7\\
CLIP &ViT-B&41.6 &65.2 & 74.8&72.9 &85.6&90.9\\

BLIP &ViT-S&44.9 &68.6 &77.4 &73.7 &86.2 &91.4\\
BLIP &ViT-B&48.2 &72.7 &80.6 &76.3 &88.6 &92.5\\

BLIP2 &ViT-S&47.3 &71.6 &80.1 &75.9 &88.1 &92.4\\
BLIP2 &ViT-B&48.8 &72.6 &81.2 &77.3 &89.4 &93.3\\
\hline
PLIP &RN50&52.9 &74.7 &82.5 &81.1 &91.4 &94.6\\
 \rowcolor{gray!20}
PLIP &Swin-B&\textbf{56.3} &\textbf{77.6} &\textbf{84.8} &\textbf{83.7} &\textbf{93.6} &\textbf{95.9}\\
\hline
\end{tabular}}
\vspace{-4mm}
\end{wraptable}
PLIP exhibits clear superiority in person representation learning compared to other general pre-training methods. To validate this, we have pre-trained a series of models on the entire SYNTH-PEDES dataset with different pre-training methods, \ie CLIP~\cite{CLIP}, BLIP~\cite{BLIP}, BLIP2~\cite{BLIP2} and PLIP. For BLIP2, we exclusively utilize its first-stage pre-training method, aligning with the default setting in the original paper designed for image-text retrieval tasks. Subsequently, we compare their direct transfer performance on downstream datasets. The results, as reported in Tab.~\ref{otherpretrain}, demonstrate that under the comparable setting of models with the roughly same parameters (ViT-Small vs ResNet50 and ViT-Base vs Swin-Base), PLIP outperforms all other pre-training methods significantly on downstream datasets. Specifically, PLIP with ReNet50 achieves 52.9\% and 81.1\% rank-1 on CUHK-PEDES and Market1501, greatly surpassing CLIP with ResNet50 by 14.8\% and 11.0\%. Moreover, with Swin-Base as the backbone, PLIP achieves the best performance, with rank-1 reaching 56.3\% and 83.7\%, respectively. These experimental results demonstrate the superiority of PLIP in person representation learning.

\subsection{More Evaluation on Downstream Tasks}

\begin{wraptable}[9]{r}{7cm}
\vspace{-6.5mm}
\centering
\caption{Comparisons of domain generalization performance on image-based person Re-ID.}
\label{generalreid}
\resizebox{\linewidth}{!}{
\begin{tabular}{c|c|cc|cc}
\hline
\multicolumn{1}{c|}{\multirow{2}{*}{Method}} & \multicolumn{1}{c|}{\multirow{2}{*}{Train Set}}&\multicolumn{2}{c|}{Market1501}&\multicolumn{2}{c}{MSMT17}\\
\cline{3-6}
& & mAP & R@1 & mAP & R@1  \\
\hline

QAConv&RandPerson &46.9 &74.5 &14.0&40.6 \\

QAConv&ClonedPerson &\textbf{59.9} &\textbf{84.5} &18.5&49.1 \\

TransMatcher&RandPerson &49.6 &77.6 &16.4&45.3 \\

TransMatcher&UnrealPerson &59.4 &81.6 &21.6&52.0 \\

WePerson&WePerson &55.0 &81.5 &18.9&46.4\\

APTM&MALS &3.8 &11.9 &1.8&7.4 \\
\hline
 \rowcolor{gray!20}
PLIP&SYNTH-PEDES &55.8 &81.1 &\textbf{22.1}&\textbf{52.3} \\
\hline
\end{tabular}}
\end{wraptable}
\textbf{Domain Generalization for Image-based Person Re-ID.} 
A comparison to the SoTA in generalizable image-based person re-identification is shown in Tab.~\ref{generalreid}. The models trained on the source datasets are directly generalized to the target datasets without tuning. Several methods and datasets recently are compared, with methods including QAConv~\cite{QAConv}, TransMatcher~\cite{TransMatcher}, WePerson~\cite{WePerson}, and APTM~\cite{APTM}, and datasets including RandPerson~\cite{RandPerson}, ClonedPerson~\cite{ClonedPerson}, UnrealPerson~\cite{UnrealPerson}, WePerson~\cite{WePerson} and MALS~\cite{APTM}. As the results indicate, even without specific design tailored for this generalizable task, our PLIP trained on SYNTH-PEDES shows competitive performance when directly generalized to downstream real-world datasets. These results verify the outstanding generalization ability of our pre-trained models for this task.

\begin{wraptable}[15]{r}{7cm}
\vspace{-4.5mm}
\centering
\caption{Improving two person attribute recognition baseline methods. The results of DeepMAR$\ddagger$ are from a re-implementation by replacing the backbone with ResNet50, which are much better than the original.}
\label{tab7}
\resizebox{\linewidth}{!}{
\begin{tabular}{c|c|cccc|cccc}
\hline
\multicolumn{1}{c|}{\multirow{2}{*}{}} & \multicolumn{1}{c|}{\multirow{2}{*}{Pre-train}}&\multicolumn{4}{c|}{PA100k}&\multicolumn{4}{c}{PETA}\\
\cline{3-10}
& & mA & Acc& Rec& F1 & mA & Acc& Rec& F1 \\
\hline
\multicolumn{1}{c|}{\multirow{6}{*}{\rotatebox{90}{DeepMar~\cite{a1}}}}
&{Baseline} &78.3&76.8 &84.0&84.3&82.7&77.1&84.6&85.2 \\
&{MoCov2~\cite{mocov2}} &78.1&76.6 &84.1&84.2&82.9&77.1&84.8&85.4\\
&{CLIP~\cite{CLIP}} &79.4&77.5 &84.9&85.2&83.3&77.6&84.8&85.4\\
&{LUP~\cite{LUP}} &78.5&77.3 &84.5&84.6&82.8&77.3&84.5&85.3\\
&{LUP-NL~\cite{LUPnl}} &79.5&77.4 &84.7&84.5&83.5&77.9&85.0&85.6\\

&{\cellcolor{gray!20}PLIP}&\cellcolor{gray!20}\textbf{80.5}&\cellcolor{gray!20}\textbf{78.8}&\cellcolor{gray!20}\textbf{86.0}&\cellcolor{gray!20}\textbf{86.5}&\cellcolor{gray!20}\textbf{83.8}&\cellcolor{gray!20}\textbf{78.2}&\cellcolor{gray!20}\textbf{85.3}&\cellcolor{gray!20}\textbf{85.8}\\
\hline
\multicolumn{1}{c|}{\multirow{6}{*}{\rotatebox{90}{Rethink~\cite{rethinking}}}}
&{Baseline} &80.2&79.2 &87.0&87.4&84.0&78.7 &85.6&86.4\\
&{MoCov2~\cite{mocov2}} &80.2&79.0 &87.0&87.3&83.4&77.7&84.9&85.6\\
&{CLIP~\cite{CLIP}} &81.1&79.8 &87.7&87.8&84.5&79.2&86.0&86.7\\
&{LUP~\cite{LUP}} &80.2&79.5 &87.3&87.6&84.2&78.3 &85.2&86.1\\
&{LUP-NL~\cite{LUPnl}} &81.3&79.1 &87.7&87.4&84.1&78.6 &85.2&86.3\\

&{\cellcolor{gray!20}PLIP} &\cellcolor{gray!20}\textbf{82.2}&\cellcolor{gray!20}\textbf{81.1} &\cellcolor{gray!20}\textbf{88.6}&\cellcolor{gray!20}\textbf{88.6}&\cellcolor{gray!20}\textbf{85.1}&\cellcolor{gray!20}\textbf{79.5}&\cellcolor{gray!20}\textbf{86.3}&\cellcolor{gray!20}\textbf{86.9}\\
\hline
\end{tabular}}
\end{wraptable}
\textbf{Improvement over Person Attribute Recognition Methods.}
Our model can also bring considerable improvement to person attribute recognition methods. We conduct experiments using two representative baseline methods~\cite{a1,rethinking} by comparing the performance gain of different pre-trained ResNet50 models. We report the results in Tab.~\ref{tab7}, where the two methods are based on traditional CNN structure and multi-label classification loss. As we can see, our model improves their performance significantly on the two popular datasets, and the improvements leave all other pre-trained models far behind. Particularly, in terms of \textit{mA}, by using our pre-trained ResNet50, the average improvements of these methods are 2.1\% and 1.1\% on PA100k and PETA respectively. These results demonstrate that our framework helps to learn better person representations for fine-grained recognition.

\begin{wraptable}[16]{r}{7cm}
\vspace{-4mm}
\tiny
\centering
\caption{Comparison with SoTA methods on human parsing. We show the best score in bold.}
\label{human_parsing_sota}
\resizebox{\linewidth}{!}{
\begin{tabular}{c|c|c|c}
\hline
\multicolumn{1}{c|}{\multirow{2}{*}{Methods}} & \multicolumn{1}{c|}{\multirow{2}{*}{Backbone}}&\multicolumn{1}{c|}{LIP}&\multicolumn{1}{c}{PASCAL}\\
\cline{3-4}
&  & mIoU  & mIoU  \\
\hline
\multicolumn{1}{c|}{Attention~\cite{HP_Attention}}&VGG16 &42.92 &56.39 \\

\multicolumn{1}{c|}{MMAN~\cite{MMAN}}&RN101 &46.81& 59.91 \\

\multicolumn{1}{c|}{JPPNet~\cite{JPPNet}}&RN101 &51.37 &59.36  \\ 

\multicolumn{1}{c|}{CE2P~\cite{CE2P}}&RN101 & 53.10&- \\

\multicolumn{1}{c|}{CNIF~\cite{CNIF}}&RN101 &57.74&70.76  \\ 

\multicolumn{1}{c|}{SCHP~\cite{SCHP}}&RN101 &59.36 &71.46 \\ 

\multicolumn{1}{c|}{CDGNet~\cite{CDGNet}}&RN101 & 60.30&- \\
\multicolumn{1}{c|}{SOLIDER~\cite{SOLIDER}}&Swin-B & 60.50&- \\

\hline
\multicolumn{1}{c|}{PLIP}&RN50 &60.41  &72.14  \\
\multicolumn{1}{c|}{PLIP}&RN101  &61.32&72.63\\
\multicolumn{1}{c|}{PLIP}&RN152&62.44  &73.51   \\
 \rowcolor{gray!20}
\multicolumn{1}{c|}{PLIP}&Swin-B&\textbf{63.52}  &\textbf{73.93}   \\
\hline
\end{tabular}}
\end{wraptable}
\textbf{Comparison with state-of-the-art methods on human parsing.} Human parsing task requires excellent perception of spatial information. By training the TIC task, as shown in the Figure~\ref{fig:sic_colorize}, our models truly learn the correlation between attribute phrases and spatial regions, which guarantees the good performance of this fine-grained semantic segmentation task. In Tab.~\ref{human_parsing_sota}, we compare our results with existing SoTA human parsing methods on LIP~\cite{LIP} and PASCAL~\cite{PASCAL}. Under the ResNet101 setting, our method achieves 61.32\% and 72.63\% mIoU on LIP and PSCAL, respectively, significantly surpassing previous SoTA methods. Also, by applying our pre-trained Swin-Base model on SCHP~\cite{SeqNet}, we achieve the best performance on each datasets. These results demonstrate that our pre-training framework shows good potential in learning discriminative person representation for this task.

\end{document}